\documentclass[preprint]{elsarticle}

\usepackage{lineno,hyperref}
\usepackage{threeparttable}
\usepackage{makecell}
\usepackage{bm}
\usepackage{xcolor}
\urldef\myurl\url{https://en.wikipedia.org/wiki/How_It%27s_Made}

\renewcommand\labelenumi{%
  {\ifcase\value{enumi}% 0
  \or\color{black}% 1
  \or\color{black}% 2
  \or\color{black}% 3
  \or\color{black}% 4
  \or\color{black}% 5
  \else\fi%
  \Roman{enumi}.}%
}
\makeatother

\setcounter{secnumdepth}{4}

\newcommand{\revision}[1]{{\color{black} #1}}

\journal{Sustainable Production and Consumption}

\bibliographystyle{elsarticle-num}
\begin{document}

\begin{frontmatter}

\title{Material Measurement Units for a Circular Economy: Foundations through a Review}

%% Group authors per affiliation:
\author{Federico Zocco\fnref{myfootnote1}\corref{mycorrespondingauthor}}
\fntext[myfootnote1]{Centre for Intelligent Autonomous Manufacturing Systems, School of Electronics, Electrical Engineering and Computer Science, Queen's University Belfast, Northern Ireland, UK; email: f.zocco@qub.ac.uk, s.mcloone@qub.ac.uk.}

\author{Se\'an McLoone\fnref{myfootnote1}}

\author{Beatrice Smyth\fnref{myfootnote2}}
\fntext[myfootnote2]{Research Centre in Sustainable Energy, School of Mechanical and Aerospace Engineering, Queen's University Belfast, Northern Ireland, UK; email: beatrice.smyth@qub.ac.uk.}

\cortext[mycorrespondingauthor]{Corresponding author}

\begin{abstract}
Long-term availability of minerals and industrial materials is a necessary condition for sustainable development as they are the constituents of any manufacturing product. To enhance the efficiency of material management, we define a computer-vision-enabled material measurement system and provide a review of works relevant to its development with particular emphasis on the foundations. A network of such systems for wide-area material stock monitoring is also covered. Finally, challenges and future research directions are discussed. As the first article bridging industrial ecology and advanced computer vision, this review is intended to support both research communities towards more sustainable manufacturing.
\end{abstract}


\end{frontmatter}

\section{\revision{Introduction}}
In contrast to the current linear economy, which can also be described as “make, take, dispose”, the circular economy aims to break the link between economic growth and consumption of finite resources \cite{EllenAndGranta2015}. In a circular economy, materials are recirculated, reduced, reused, recycled and recovered, and the overall goal is “to keep products, components and materials at their highest utility and value, at all times” \cite{EllenAndGranta2015}. Key benefits of a circular economy are the contribution to environmental protection, specifically, within an EU context, in terms of carbon and biodiversity targets, which are at the core of the European Green Deal \cite{EuropeanCommission2018,EuropeanCommission2019,EuropeanCommission2020} and its ambition to move to a clean, circular and sustainable economy. 

A circular economy could also bring benefits in terms of jobs. A recent EU study showed that a net increase in jobs and GDP of 700,000 and 0.5\%, respectively, could be achieved by 2030 through moving to a more circular economy \cite{CambridgeEconometrics2018}. Such a transition would support the United Nations Sustainable Development Goals (SDGs), including SDG11 Sustainable Cities and Communities, SDG12 Responsible Production and Consumption, and SDG13 Climate Action \cite{UnitedNations2030Agenda}. However, despite the obvious benefits, there is still much progress to be made. The EU produced 2.6 billion tonnes of waste in 2016, while in 2017 almost one quarter of waste (24\%) in the EU-28 was sent to landfill; values vary across member states, with five states landfilling over 70\% of municipal waste \cite{EuropeanParliamentWaste2021}. The OECD has predicted that global materials use will be more than double the 2011 value by 2060 \cite{OECDoutlook2018}, while global annual waste generation is expected to increase by 70\% by 2050 \cite{kaza2018waste}.

A key question is therefore: how do we move from our current linear economy to a circular economy for materials? Initiatives are needed on many fronts, but central to the transition is measurement and tracking, as we cannot manage what is not measured. Consider the water network as an analogy. Clean water can be tracked from extraction, through the treatment stages and distribution network, and finally at the end user. Wastewater can also be measured as it travels in the network to the treatment works before final sludge disposal and release of effluent. Measurement of water throughout its cycle can have a positive impact on resource efficiency, as leaks and hotspots can be identified and better management practices developed \cite{adedeji2017leakage}. Like water, materials are a state of matter. The hypothesis arises that if we can track water and use this information to improve resource efficiency, then we can also track materials through their life cycle, from material extraction to manufacturing, retail, the consumer and finally to waste, reuse or recycling.

Materials, however, are heterogeneous, do not flow through pipes like water and so cannot be measured and tracked in the same way. New systems are therefore required, but existing research on the topic is limited. Previous reviews of automation for materials have focused on waste management, with only two such reviews found by the authors. To aid the planning and design of sustainable new systems, Hannan et al. \cite{hannan2015review} reviewed technologies in solid waste monitoring and management systems and developed classifications to enable selection of ICTs for a particular problem. Gundupalli et al. \cite{gundupalli2017review} reviewed automated sorting of source-separated municipal solid waste for recycling with a view to supporting system designers and identifying areas for future research. The paper concluded that there is a need for robotic systems to deal with mixed wastes in large landfill sites in developing countries. The paper also highlighted that traditional person-based managing of wastes is highly labour intensive and can be dangerous and expensive.

These previous reviews made interesting contributions, but focused only on technologies for the waste sector and did not consider the holistic life cycle of materials as is required for a circular economy, which is about more than just waste. A further gap in previous research is the lack of detailed exploration of the potential of computer vision to support the transition to a circular economy. With the overall goal of improving the management of natural resources, the aim of this paper is to provide a guide for the design of a material monitoring sensor network enabled by computer vision that could be used across the entire manufacturing life-cycle, from raw material extraction to manufacturing, retail, consumers, waste sorting, recycling and landfill. 

\textbf{Our contributions.} The main contributions of this review are the following. 
\begin{enumerate}
\item{We define a computer-vision-enabled system to monitor stocks of materials and we provide a review of works relevant to its development with particular emphasis on the foundations. Then, we provide a guide towards the design of a material monitoring sensor network.}
\item{This is the first article, review or otherwise, that bridges the gap between computer vision and the holistic perspective on manufacturing supply chains that is at the core of industrial ecology and circular economy. Indeed, as we will show, while previous works have proposed vision systems for waste management or material recognition, we differ from the former by looking at manufacturing materials regardless of their life-cycle stage and from the latter by putting material recognition within the context of natural resource management.}
\item{We cover the latest advances in computer vision, from the theory to implementation aspects and related challenges.}
\item{Finally, we delineate five research directions which may inspire both computer engineers with environmental concerns and industrial ecologists or circular economists interested in exploiting the latest advances in computer vision.}
\end{enumerate}      

The paper is organized as follows. \revision{Section \ref{sec:Background} covers the background literature on computer vision, Section \ref{sec:Methods} explains the methodology used to select the references included in this review which is provided in Section \ref{sec:Results}; then Section \ref{sec:Challenges} summarizes the outcome of this review and discusses the main challenges and future research directions; finally Section \ref{sec:Conclusions} concludes.}

\section{\revision{Background Literature on Computer Vision}}\label{sec:Background}
%Here make an overview of computer vision with emphasis on deep learning. In this way the reader is ready to understand the next section which goes into the MMU.
Computer vision is a research area concerned with making useful decisions about real physical objects and scenes based on sensed images \cite{shapiro2000computer}. It is a subfield of artificial intelligence and consists of designing a \emph{signal-to-symbol converter} \cite{minsky1961steps}: cameras provide signals (i.e. measurements) about the physical world and the computer vision model converts them into symbolic representations, e.g. the word/symbol ``cat'' if the image depicts a cat.  

Research in computer vision spans more than forty years \cite{szeliski2020computer}. State-of-the-art techniques can be divided into two broad categories: hand-designed feature methods (i.e. classical machine learning) and representation learning methods (i.e. deep learning). For the foundations of the field, the reader should refer to well-established books such as \cite{szeliski2020computer} for an updated and general treatment, \cite{forsyth2012computer} for a focus on the first category of techniques and \cite{Goodfellow-et-al-2016,zhang2020dive} for a focus on the second category. This section provides the general concepts of the two broad categories.

\begin{figure}
\begin{minipage}{\textwidth}
\centering
\includegraphics[trim={0 5cm 0 0},width=0.85\textwidth]{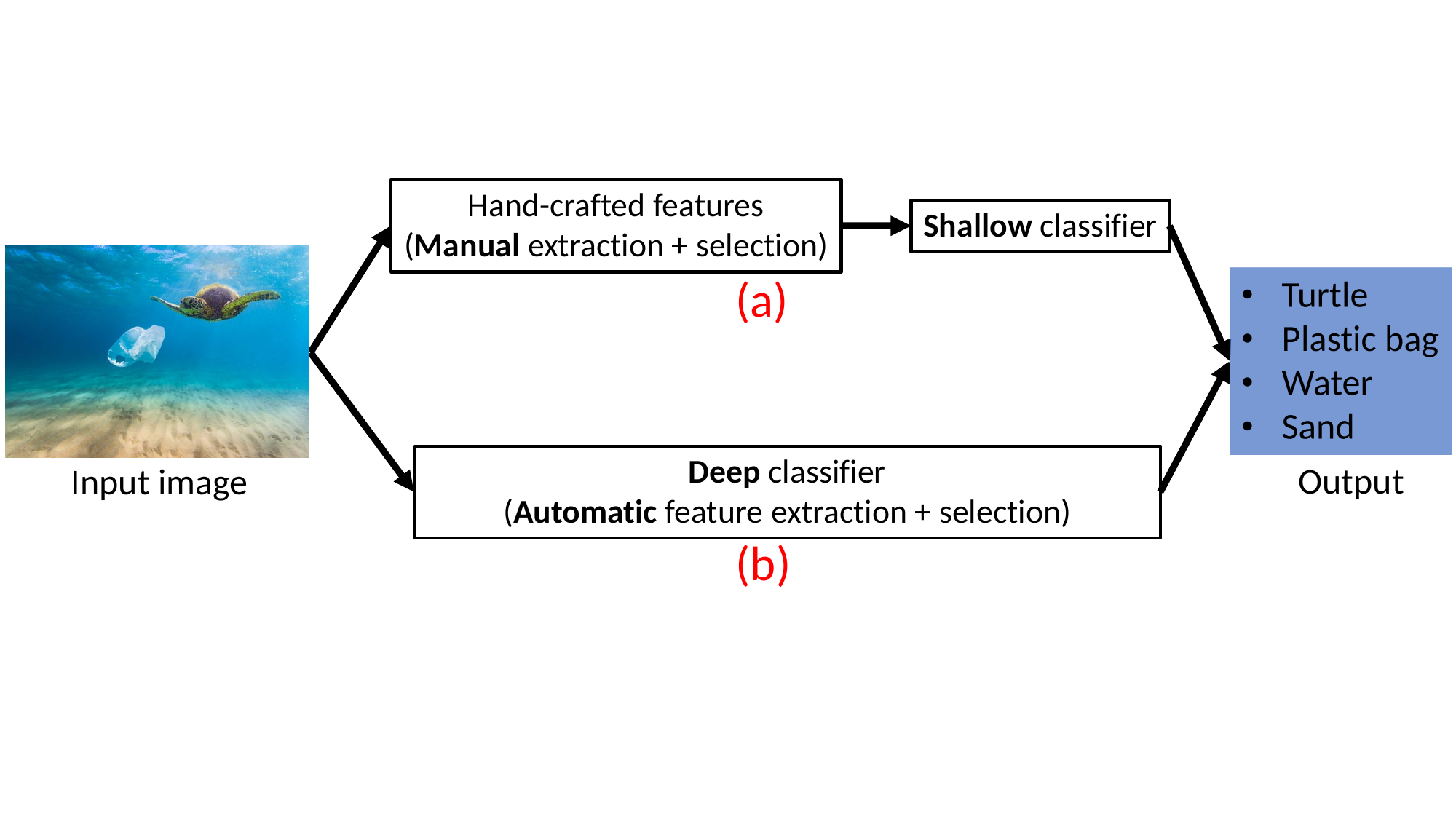}
\caption[...]{Main difference between (a) hand-crafted feature methods and (b) representation learning methods\textsuperscript{a}.}\tiny\textsuperscript{a}Turtle image sourced from: \url{https://www.pinterest.co.uk/pin/plastic-pollution-in-the-news-and-why-we-should-care-ukonserve--278026976982870528/}
\label{fig:ClassicVSmodernCV_cropped}
\end{minipage}
\end{figure}

The main difference between hand-crafted feature and representation learning methods is depicted in Fig. \ref{fig:ClassicVSmodernCV_cropped}: the former requires an expert to design the algorithm/model capturing the characteristic features of the image of interest, the latter instead assumes that the computer learns them during a training phase; as a consequence, the former requires expert domain knowledge and less computing resources than the latter. Given that a deep learning of representations works directly between the input images and the output symbols, the second category is also referred to as the end-to-end learning approach \cite{wang2018deep}.

% The first category is split in two: (1.1) techniques for features extraction and (1.2) for classification; (2) the second category is all in one part.
% for (1.2) see Section 5.1 and 5.2 of ``UpdatedCVBook''; for (1.1) see Section 6.2.1 of the same book; for (2) see ch. 6 of ``NewBookOnDL'' and Table I of paper ``CNNvariantsForClassification''. 
\noindent
\subsection{Hand-Crafted Feature Methods}  
One of the simplest methods is based on bag-of-words (BoW) \cite{szeliski2020computer}, also called bag-of-features, which identifies a set of keywords (i.e. features) for each image analogously to text documents that are described by the word content. Then, a test image is assigned to the class with the closest word/feature composition. One of the first BoW recognition systems was proposed in \cite{csurka2004visual} (see also \cite{csurka2006generic}): the image patches are detected using Harris affine \emph{detectors} \cite{mikolajczyk2002affine}, which in turn are used to compute the scale invariant feature transform (SIFT) \emph{descriptors} \cite{lowe1999object}; then, a histogram of visual words is used as the input vector to a machine learning classifier. Other types of detectors and descriptors are compared in \cite{zhang2007local}.

Bag-of-words models are the simplest because they do not consider the geometric relationships between different parts and features \cite{szeliski2020computer}. While this makes them particularly efficient, higher inference accuracy is provided by \emph{part-based} models which focus on the geometric relationships between the constituent parts of the object \cite{felzenszwalb2005pictorial}. More details on part-based modeling can be found in \cite{FergusCourse2009}.  

An approach even more accurate than part-based modeling takes into account the \emph{context} in which the object with its constituent parts occur \cite{oliva2007role}. Combinations of part-based and context models in the same vision system have also been proposed \cite{sudderth2008describing,crandall2007composite}.
    
% Now classical machine learning for the classifier: 
% First method to define the classifier:    
As visible in Fig. \ref{fig:ClassicVSmodernCV_cropped}, the final block in the pipeline is the classifier. One of the simplest classification algorithms is the \emph{$k$-nearest neighbors} which consists of finding the $k$ training samples closest to the new sample and evaluating its class knowing the class of the neighbors \cite{szeliski2020computer}. A library with nearest neighbors algorithms for large training datasets is presented in \cite{muja2014scalable}. 

% Second method to define the classifier:
The $k$-nearest neighbors is a non-parametric approach since it does not define a model of learned parameters from the training set. A simple parametric classification algorithm is \emph{multiclass logistic regression} (despite the name, this method is not for regression), which learns a linear model and applies the \emph{softmax} function to the model output to give the probability of having the class $C^i$ given the input feature vector $\bm{x}$, i.e. $p(C^i|\bm{x})$ \cite{bishop2006pattern}. 

%Third method to define the classifier:
In some cases there are multiple possible surfaces that correctly divide the training samples into their classes. In these cases kernel \emph{support vector machines} (SVMs) define the decision boundary as the one that maximizes the distance between the training set classes \cite{bishop2006pattern}. A review of kernel-based methods for computer vision can be found in \cite{lampert2009kernel}.

% Fourth method to define the classifier:
Another approach consists of using \emph{decision trees} having a graph structure. The key idea of this approach is to divide the complex classification task into simpler tests that are hierarchically organized \cite{szeliski2020computer}. For example, assume we have an image of an outdoor garden; to classify it as ``outdoor garden'' the problem can be split in two subsequent steps: the first answering ``Is there sky at the top?'' and, if true, the second answers ``Is the bottom part green?'' \cite{criminisi2013decision}.

\subsection{Representation Learning Methods} 
The most used computer vision methods belonging to this second category are based on \emph{convolutional neural networks} (CNNs) \cite{lecun2010convolutional,zhang2020dive}. CNNs consist of a network architecture designed to perform computations emulating multiple connected layers of neurons in a fashion similar to the neural network of a human brain. \revision{Most CNN-based computer vision systems require five components: the data, the neural model, the training algorithm, the cost function (aka ``loss function'') and the performance metric \cite{Goodfellow-et-al-2016}.
\begin{enumerate}
\item{\textbf{Data.} Representation learning methods are algorithms that generate a nonlinear map between input and output data. The data contain the information about the task being addressed and the neural model has to be trained to fit the specific data. For example, if we need a vision system that recognizes the digits from 0 to 9, the input data are the images of the digits and the output is an integer that uniquely identifies the type/class of an input image (see \cite{MATLABtutorial} for a MATLAB tutorial). A vision algorithm sees each image as a matrix whose elements specify the position and the color of each pixel. If the neural model is trained properly, it will classify the images of 0-9 digits with high accuracy. This is analogous to the way we learned such a classification: when we were born, an image depicting a zero had no meaning; subsequently, at school, we have been trained to associate the picture of a digit with a quantity, e.g. zero apples, zero toys.}
\item{\textbf{Neural model.} A neural model is an artificial emulator of a biological neural network. A single artificial neuron performs a simple linear combination between the input signal and the network \emph{weights} and then applies a nonlinear function (called ``activation function'') that produces a nonlinear mapping between the input signal and the neuron output. While the model of a single artificial neuron is simple, complex models can be achieved when multiple neurons are connected in a neural network. A neural network is a stack of multiple layers and each layer is a set of neurons. A shallow neural network becomes a \emph{deep} one by stacking multiple layers. The weights define the intensities of the connections between the neurons. The architecture of the neural model is designed before training and the weights are usually initialized with random values. Subsequently, the values of the \emph{weights} change during training in order to adapt the neural network to the data. The training algorithm discussed below decides how to change the weights during training in order to fit the data.}
\item{\textbf{Training algorithm.} The training algorithm (aka ``optimizer'') defines how to update the \emph{weights} during the training taking into account the data along with the loss function being optimized. Hence, it \emph{optimizes} the network weights for the specific task. In general, the best training algorithm is the one that provides the most optimal weight values in the shortest time. For an introduction to the different types of training algorithm that have been developed, the reader could refer to \cite{Goodfellow-et-al-2016}, Chapter 8.}
\item{\textbf{Loss function.} The loss function is a mathematical description of the goal of the neural model. Recalling the example of digit classification mentioned above, the loss function employed is \emph{cross-entropy} (detailed in \cite{bishop2006pattern}, Chapter 4) which measures the classification error committed by the model during training. Therefore, minimizing the cross-entropy function, in practice, means minimizing the classification error of the model. Since loss functions usually have multiple local minima, there is no guarantee that the optimizer will find the global optimum.}
\item{\textbf{Performance metric.} Once the training is complete, the weights of the network are adapted to the data according to the loss function. The performance metric is used to evaluate the accuracy of the trained model. Recalling the example of the digits, a performance metric could be the percentage of correct classifications made by the model when processing a sequence of digit images. The performance metric changes if the task differs from classification, e.g. mean squared error for regression problems or average precision in object detection \cite{ren2015faster}.}
\end{enumerate}
} 

\begin{figure}
\centering
\includegraphics[trim={0 0 0 7cm}, width=0.9\textwidth]{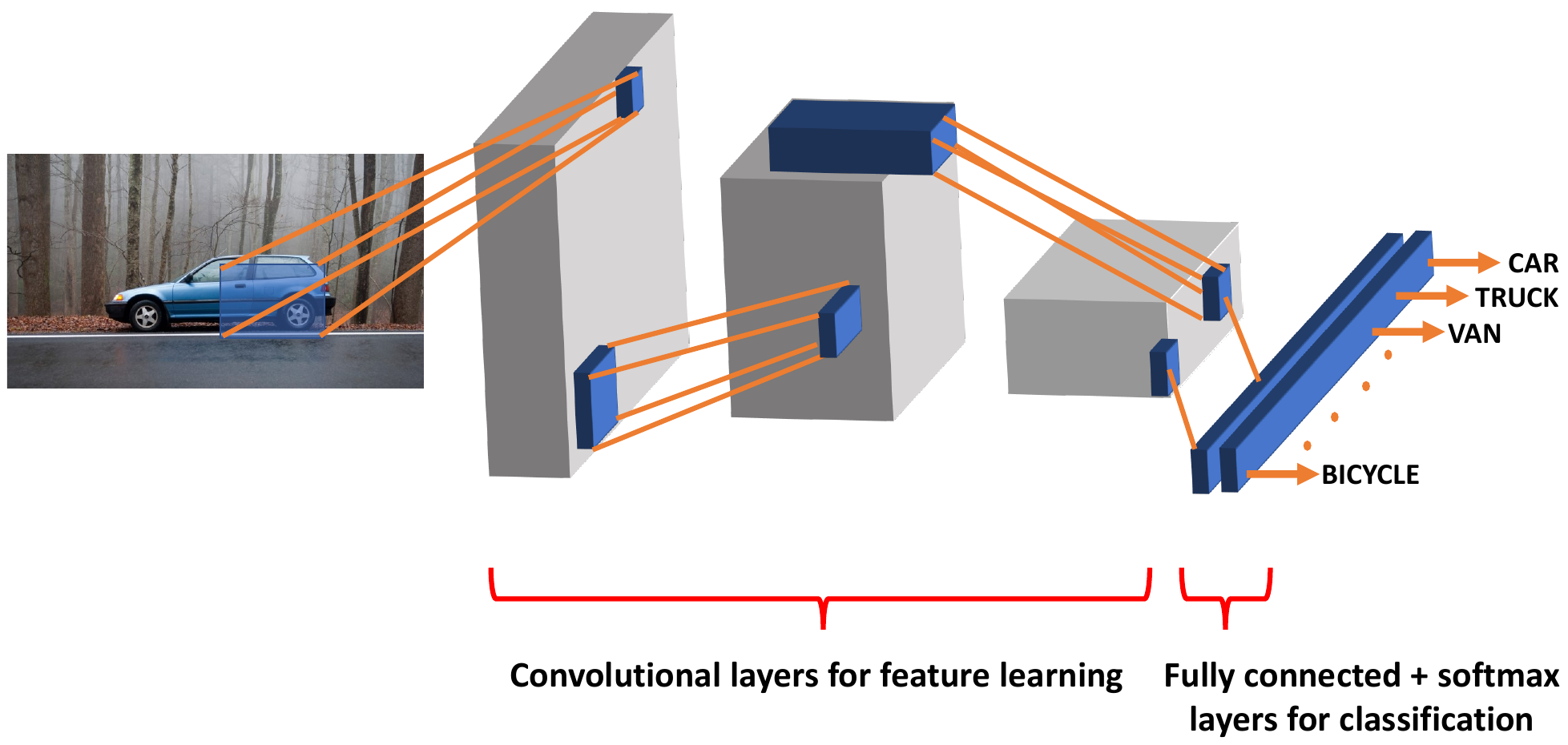}
\caption{\revision{Example of the architecture of a simple convolutional neural network: from the input image the features are learned with hierarchical levels of abstraction using multiple convolutional and other types of layers such as ReLU and pooling (gray blocks); those operations transfer the local information of a layer into the next (blue blocks). Finally, the softmax layer converts the features into classes processing the output of one or more fully-connected layers (last blue blocks at the end of the pipeline) \cite{Goodfellow-et-al-2016}.}}
\label{fig:CNNarchitecture_cropped}
\end{figure}

By definition, a CNN has at least one neural layer that performs the \emph{convolution} operation \cite{Goodfellow-et-al-2016}. An example of a CNN architecture is depicted in Fig. \ref{fig:CNNarchitecture_cropped}: training images are given as input and, once the end-to-end learning is completed, the resulting model gives the conditional probabilities $p(C^i|\bm{x})$, where $C^i$ is the $i$-th class (e.g. car, bicycle) and $\bm{x}$ is the input image. Therefore, the features are embedded into the model parameters defined through the optimization of the cost function. A key reason that has made CNNs particularly successful for computer vision over other neural network architectures is that they accept a two-dimensional input and, through convolutions, perform two-dimensional operations. Hence the pixels of the input image are processed preserving their original relative position. 

% R-CNN
Different CNN architectures have been proposed over the years \cite{zhao2019object}. For example, in \cite{girshick2014rich} firstly several local regions of the input image are identified, secondly a large CNN learns the features of each local region and finally it classifies the content of each region using a linear SVM per class. Subsequently this CNN architecture has been sped-up in \cite{girshick2015fast,ren2015faster}.     

% SPP-net
The basic CNN can process images of arbitrary size with the output size of the convolutional layers (also called \emph{feature maps}) being influenced by the image input size. As a consequence, a trained CNN may have an architecture which is suitable for processing images of a fixed size (say 256 $\times$ 256), but may not be adaptable to process images of a different size (say 128 $\times$ 128). Therefore, \cite{he2015spatial} proposes placing a layer after the last convolutional layer in order to make it possible for the network to process different image sizes.  

% FCN, then R-FCN
As discussed for Fig. \ref{fig:CNNarchitecture_cropped}, typically the output of the model is the marginal probability $p(C^i|\bm{x})$. The \emph{fully convolutional} network in \cite{long2015fully} gives instead such a probability pixel-to-pixel, i.e. the output is a two-dimensional matrix giving the class of each pixel. Subsequently, the region-based classification approach cited above (i.e. \cite{girshick2014rich}) has been combined in \cite{dai2016r} with a fully convolutional network.     

% YOLO, then SDD
A more complete object classification system is proposed in \cite{redmon2016you}, where for a given input image depicting a scene, the model infers both the class of every detected object and a bounding block around them to locate their spatial position. Specifically, the model divides the input image into a grid and, stating the problem as a regression task, for each grid cell it predicts multiple bounding boxes, the confidence for those boxes and the object class probabilities. The authors initially define a smaller CNN network and then, once pre-trained, they convert the model to perform detection adding four convolutional layers, two fully connected layers and increasing the input resolution of the network from 224 $\times$ 224 to $448 \times 448$. Subsequently, this algorithm has been sped-up and made more accurate in \cite{liu2016ssd}.      

For a more comprehensive treatment of CNNs the reader should refer to \cite{zhang2020dive,Goodfellow-et-al-2016} for the basics, \cite{zhao2019object} for a review of several variants and \cite{voulodimos2018deep} for a brief review of the most popular deep learning algorithms for computer vision (not just CNNs).

\section{\revision{Methods}}\label{sec:Methods}
\begin{figure}
\centering
\includegraphics[width=\textwidth]{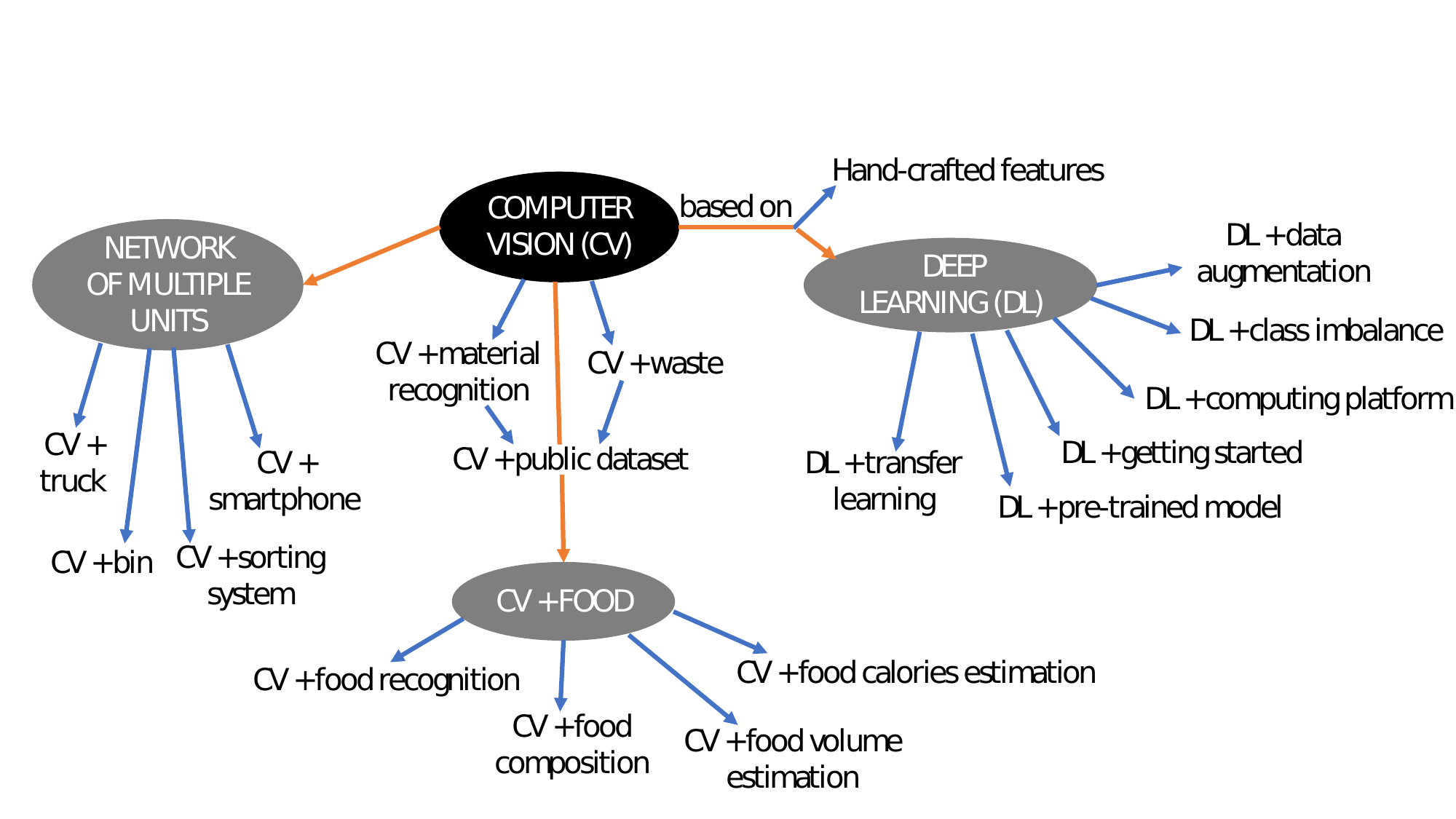}
\caption{\revision{A map of the keywords/concepts that has been followed to select the references included in this review. The arrows in orange and in blue connect the primary and the secondary concepts/keywords, respectively. Computer vision is the basis for the proposed technology, hence it has a central position and, directly or indirectly, it is connected with all the other keywords.}}
\label{fig:ProcedureToReview}
\end{figure}

\revision{This review seeks to gather the relevant literature for researching and developing the computer-vision-based technology detailed in the next section. The references were selected by following the map of keywords/concepts shown in Fig. \ref{fig:ProcedureToReview}, where Computer Vision has a central position since it defines the basis. The orange and blue arrows connect the primary and secondary concepts/keywords used in this review, respectively. Computer Vision is connected to Deep Learning on the right side since we emphasize vision algorithms based on representation learning rather than on hand-crafted features. At the bottom, it is connected to ``CV + food'' because food is a particular type of manufacturing product, hence works focused on both computer vision and food can be adapted to focus on other manufacturing products (see next section for details). On the left side, Computer Vision is connected to ``Network of multiple units'' since a key idea discussed in the next section is to design a network of multiple material measurement devices sending data to a central database; single units could be implemented on trucks, bins, smartphones, sorting plants or industrial robots. The final aim of such a network is to improve mapping and quantification of materials in a target area.}    

\section{\revision{Results}}\label{sec:Results}
\subsection{Material Measurement Unit}\label{sec:MMUdefinition}
\begin{figure*}[t]
\centering
\includegraphics[width=\textwidth]{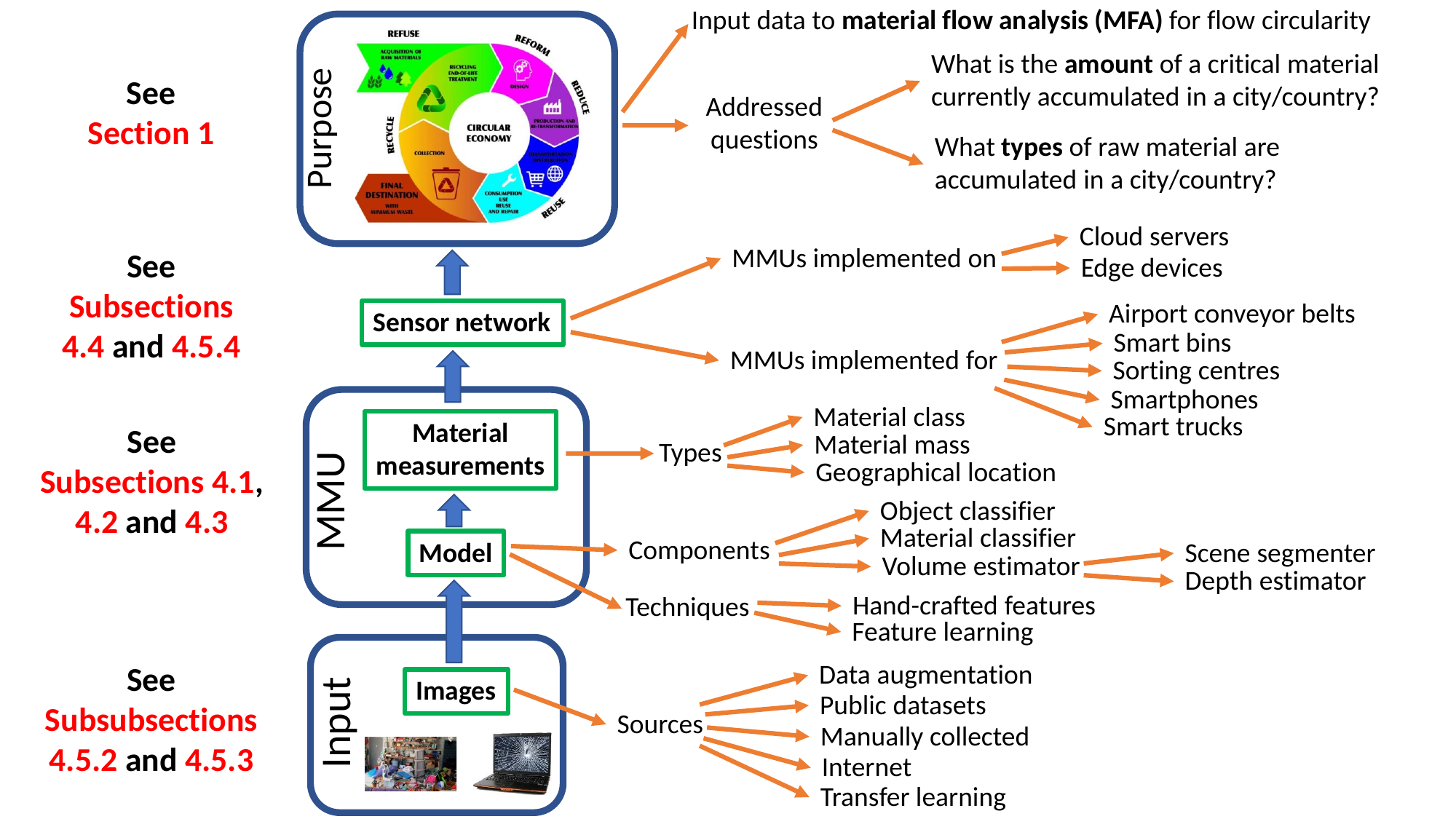}
\caption[...]{\revision{Overview of the MMU system and its purpose: an MMU receives images as input and seeks to measure material stocks in a desired location. Multiple MMUs could be connected as a sensor network and provide real-time material mapping and quantification data to MFA studies. The corresponding sections of the review are indicated along the left side\textsuperscript{a}.}}\tiny\textsuperscript{a}Circular economy image sourced from: \url{https://www.portoprotocol.com/circular-economy-as-a-way-of-increasing-efficiency-in-organizations/}
\label{fig:OverviewScheme}
\end{figure*}

The monitoring system is called a Material Measurement Unit (MMU) and is defined as follows. \\ 
\fbox{\begin{minipage}{33.9em}
\textbf{Definition 1.} \emph{An MMU is a complex sensor that, through a \revision{computer vision algorithm}, receives images of objects as input (e.g. RGB images, X-ray images, depth images) and provides as output information about the material composition of the object. The fundamental output measurements are (1) the class of material and (2) the mass of material.}
\end{minipage}}
\vspace{0.01in}
\noindent
Essentially, an MMU is a converter from object images to material measurements such as the class and the mass. Figure \ref{fig:OverviewScheme} provides an overview of the MMU context and purpose: the inputs to the system are ``Images'' as in the green rectangle at the bottom. These are collected or generated through the ``Sources'' indicated by the orange arrows on the right side; the images are processed by the MMU internal ``Model'' (i.e. the second green rectangle); the outputs of the model are ``Material measurements'' having the ``Types'' specified by the orange arrows; the fourth green rectangle mentions a ``Sensor network'', which is realized if multiple MMUs are implemented on different platforms and interconnected; the final ``Purpose'' of such a sensor network is monitoring the material stocks and flows for more sustainable natural resources management. \revision{For example, such a technology could provide further input data to material flow analysis (MFA) studies \cite{mehta2022using,millette2019materials,MFA-Europe} or extend existing material flows and stocks databases such as the Yale stocks and flows database (YSTAFDB) \cite{myers2019ystafdb}. Summarizing, the whole pipeline in Fig. \ref{fig:OverviewScheme} reads as follows from the bottom to the top: an MMU processes single images or video frames recorded through cameras to determine the \emph{type} and the \emph{mass} of materials depicted on the images or frames by using a computer vision algorithm; multiple MMUs could be deployed in different locations; the material locations could be provided by the GPS of the device implementing the MMU; the final goal of this distributed network of units is providing real-time material mapping and quantification data for MFA studies to improve the material flow circularity of a target urban area.}

An MMU can be seen as made up of three components, each one dedicated to a specific task:  
\begin{enumerate}
\item{Component 1 -- material recognition}
\item{Component 2 -- object recognition}
\item{Component 3 -- volume or mass estimation.}
\end{enumerate}
Given that the density of a material is typically a known parameter \cite{DensityOfMaterials}, the volume estimation of a product evaluated by Component 3 permits estimation of the mass through $mass$ = $density$ $\times$ $volume$. When processing an object with hidden parts such as a mobile phone, Component 1 would recognize the plastic of the case and the glass of the screen, but not the internal electronics, whereas Component 2 could recognize the specific model of a phone and read the corresponding full list of materials from a database. If instead waste plastic packaging is being processed, the complex shape of the damaged packaging makes Component 1 preferred to Component 2 because it focuses on the texture of plastic ignoring the unpredictable shape of the damaged object.     

The \revision{next three subsections are ordered considering the scale-up of the system: Subsection \ref{sub:OnMaterials} focuses on Component 1 covering works from the computer vision literature on material recognition; Subsection \ref{sub:OnFood} adds Components 2 and 3; finally Subsection \ref{sub:SensorNetwork}} proposes a distributed monitoring system exploiting multiple MMUs. Details on the topics covered by each subsection are captured on the left side of Fig. \ref{fig:OverviewScheme}.

\subsection{Computer Vision Focusing on Materials or Waste}\label{sub:OnMaterials}     
\revision{To date, typical applications of material recognition in robotics and manufacturing are in the enhancement of robotic grasping of objects or product visual quality assessment \cite{sharan2013recognizing,bell2015material}. Instead, within the context of resources management for material flow circularity, an autonomous or semi-autonomous recognition of the types of materials as in Fig. \ref{fig:OverviewScheme} could improve the mapping of resources by answering the question ``what types of materials are in a target area?''.} One of the three components of an MMU is the material recognizer, hence here we discuss previous works on computer vision for material or waste recognition.  

In \cite{schwartz2019recognizing} CNNs are trained to recognize the traits of materials using weakly-supervised learning. Particularly relevant is the result that their system is able to segment the scene with masks purely based on the material appearance, which is a local attribute, hence it does not rely on the particular shape of the objects. As noted previously, this can be useful when it comes to automatically sorting trash because products thrown away have a non-standard shape caused by the damage they experience, e.g. an empty plastic bottle could be compressed to save space, a glass bottle could be broken.

Hand-crafted-feature-based material recognition is proposed in \cite{sharan2013recognizing}, then both a generative and a discriminative model are trained from the extracted features. Moreover, to prevent the overfitting of the generative one, a greedy algorithm \cite{zocco2021greedy} is designed to add one feature at a time as long as the recognition rate increases. A conclusive suggestion of Sharan et al. \cite{sharan2013recognizing} is that the system accuracy could benefit from including the modeling of non-local features, e.g. object shape, correlated with the local surface appearances.    

The authors of \cite{wang2020multi} focus on ``waste in the wild'' proposing a two-stage scene segmentation to yield a binary waste detection system, i.e. whether there is waste in the scene or not. At the first stage, the full scene is segmented; at the second stage, a zoom-in is performed around the detected waste and the zoomed image is processed for a fine segmentation of the waste shape. The two-stage approach shows an improvement when compared to single-stage segmentations using the same neural models. For example, an MMU could use the accurate segmentation of the second stage as input to a material recognition or volume estimation system to return the type of material or estimate its mass. The authors of \cite{wang2020multi} emphasize that the task they have considered is binary, i.e. waste or not waste, because of the lack of available images for several classes of waste types (aka \emph{class imbalance}, an issue in computer vision covered in \revision{Subsubsection \ref{subsec:ClassImb}}) that makes training a segmentation system to recognize the waste type currently unfeasible. A similar problem was experienced in \cite{awe2017smart}, where to address the issue a \emph{data augmentation through trash simulation} is proposed: given a set of images of objects taken from the trash, i.e. pieces of trash, a new image is generated randomly combining the initial pieces of trash (say 2-6), thus simulating images taken from the trash rather than images of single pieces. Another possible approach for data augmentation is to generate new virtual images using rendering techniques as in \cite{li2012recognizing} or designing a platform with public access that, through crowdsourcing, grows over time as in \cite{taco2020}.

Recognizing materials from images depicting daily scenes is more challenging than with images taken in a laboratory because ``wild'' scenarios show an extremely large variation in lighting conditions, surface textures and perspectives. Hence, the authors of \cite{bell2015material} proposed a CNN-based vision system to recognize and segment the materials of images taken from the wild achieving a 73.1\% mean class accuracy. Moreover, it is shown that a fine-tuned AlexNet achieved an accuracy about 8\% higher than the hand-crafted method proposed in \cite{cimpoi2014describing} when compared for the dataset they collected for the study. A key conclusion of that work is that having a rich, well-collected, well-segmented and labeled dataset is necessary for accurate material recognition, especially in challenging environments such as real-world scenes.          

Orientation histograms such as scale-invariant feature transform (SIFT) \cite{lowe2004distinctive} and histograms of oriented gradients (HOG) \cite{dalal2005histograms} are the most commonly used low-level features for object recognition. Exploiting a kernel view the authors of \cite{bo2010kernel} generalize the definition of such low-level features and give insights on how to define novel variants. Successively, these kernel descriptors have been used in \cite{hu2011toward} for both material recognition and object recognition to investigate whether these two recognition tasks are related or not; in particular the study found that using the outputs of an object recognizer improves the material recognizer accuracy, whereas the material recognizer does not help object recognition.      

In \cite{liu2010exploring} high-level material categories are learned based on low-level and mid-level features specifically designed for material recognition. They introduce a set of mid-level features to capture the shape, the reflectance, the micro-texture aspect and the color from the image. Using all these features may cause overfitting of the training set and it is not known a-priori which features are the most relevant for material recognition. Therefore, an augmented version of the latent Dirichlet allocation algorithm \cite{blei2003latent} is developed by the authors to perform a greedy feature selection. Finally the selected features are combined together to build a material recognition system.

In \cite{varma2008statistical} material classification is performed comparing two modeling approaches: the Varma-Zisserman's classifier \cite{liu2010exploring,varma2005statistical}, that uses a bank of filters to process the image patches, and the so called ``Joint'' classifier, that directly uses the source image patches instead of the filter responses generated by filtering them. The empirical comparison suggested that the ``Joint'' classifier is more accurate. A comparison in terms of computational time/complexity is not considered in \cite{varma2008statistical}, but in the context of MMUs it is an important performance metric because small platforms such as phones or microcontrollers may impose computational limits. 

In \cite{chu2018multilayer} a CNN extracts the features by processing images of objects acquired by a camera, while another sensor measures the object weight and a third one determines whether the object is made of metal or not. Then, a neural network reads the CNN image features and other engineered features to decide if the waste item is recyclable. The dataset collected for this study is available upon request to the corresponding author. The authors of \cite{bircanouglu2018recyclenet} based their waste recognition system on CNNs and also provided an extensive performance comparison of different architectures and training strategies (e.g. pre-training, transfer learning) both in terms of accuracy and computational time. They finally proposed a reduced-complexity CNN optimized to classify some waste items. A CNN optimized for waste classification is also proposed in \cite{mao2021recycling}, while the authors of \cite{vo2019novel} used deep transfer learning.

The large production of waste electrical and electronic equipment (WEEE) with only about 20\% recycling rates and the environmental problems caused by a destroy-then-melt approach for WEEE material recovery have motivated Jahanian et al. \cite{jahanian2019see} to develop a vision system for autonomous e-waste disassembling tasks. In particular, they considered the disassembling of smartphone circuit boards and they built their neural model using the Mask R-CNN \cite{he2017mask} as the baseline. From a computer vision point of view, the problem is interesting because it requires the recognition of components with different geometries and sizes within a small space, e.g. the battery and the screws, while from a circular economy perspective e-waste is of primary concern as it contains significant fractions of critical raw materials \cite{WEEEandCRMs}.      

Motivated by the negative impact of the contamination of waste items on recycling rates, Ibrahim et al. \cite{ibrahim2019contaminet} focused on detecting the contaminating material found in municipal solid waste. This work was a collaboration between a waste management company, an automation company and a university which led to the development of a CNN-based vision system trained with over 30,000 images labeled by the waste management company staff experts. Manually labeling a large number of samples is a tedious process, hence, in some cases, only unlabeled images are available and, therefore, supervised methods such as CNN are not suitable. In this case, the model proposed in \cite{lagunas2019similarity}, which proposes a similarity measure for material appearance, could be applied to cluster the unlabeled samples.\\
\fbox{\begin{minipage}{33.9em}
\revision{\textbf{Subsection take-aways:} several works on material recognition have been proposed seeking to improve robotic grasping; both hand-crafted-feature-based and CNN-based methods were proposed; waste detection in a laboratory setting is easier than``in the wild''; object recognition helps material recognition.}
\end{minipage}}

\subsection{Computer Vision Focusing on Food}\label{sub:OnFood}
Food computing is the research area seeking to make machines able to process food images and extract information such as the type of food in the image, whether there is food or not in the image, how much food is in the image, what is its recipe and how many calories it contains \cite{min2019survey}. Such information helps the machine user (e.g. the user of a mobile phone) to monitor his/her diet and modify the diet for the benefit of his/her health \cite{min2019survey}. We observe that \emph{food is an object, i.e. a manufactured product, made of organic materials}, hence research questions and challenges addressed in food computing are closely related to material computing. As we show in this section, the reader can get useful insights from food computing for developing MMUs by simply looking at food as an object, at the recipe as the object material composition, and at the food portion volume estimation as the object component volume estimation. \revision{The final goal here remains as depicted in Fig. \ref{fig:OverviewScheme}: automation or semi-automation of resources mapping and quantification; a network of MMUs could provide input data of MFA studies to design circular supply chains \cite{mehta2022using,millette2019materials,MFA-Europe} or extend existing material flows and stocks databases such as the YSTAFDB \cite{myers2019ystafdb}.}

\subsubsection{From food to material recognition} 
For example, the food recognition approach in \cite{wu2009fast} uses SIFT descriptors \cite{lowe2004distinctive} to compute the most likely food types appearing in the frames of a video recording a volunteer eating in a restaurant. Similarly, considering that unused or faulty objects accumulated in private houses might be a valuable source of materials, an MMU could process a video of these objects provided by the household to estimate the type and mass of each detected material. 

In \cite{anthimopoulos2014food} a SIFT-based bag-of-features model \cite{o2011introduction} followed by an SVM classifier is designed after an extensive investigation considering a dataset of 5000 food images organized in 11 classes. The final classification accuracy, of the order of 78\%, could be similarly achieved optimizing the model to process images taken from trash, which are complex to classify because the objects frequently have a different shape once thrown away (e.g. a bottle deformed to save space, a package that has been damaged to extract the contents). 

The authors of \cite{yang2010food} use descriptors based on the relative geometric position of the ingredients exploiting the fact that a type of food has ingredients arranged in predictable spatial configurations, e.g. a sandwich has ingredients distributed linearly over multiple layers, a plate of salad has ingredients distributed horizontally all over the plate. A similar modeling approach could effectively exploit the predictable relative position of the components of a manufacturing product, e.g. an electrical machine is composed of a rotor inside a stator, a phone is externally composed of a screen on top of a case, books are multiple layers of sheets.

In \cite{kawano2015foodcam} the focus was primarily on computationally simple detection models because it was intended to be implemented on mobile phones for a real-time food recognition application. The authors describe how the application works and its performance considering two implementations: the first one uses a bag-of-features model with SURF features \cite{bay2008speeded}, the second uses a Fischer vector model \cite{perronnin2010improving} with HOG \cite{dalal2005histograms}; both use the extracted image features as input for a linear SVM classifier. Similarly, a mobile phone application for material measurement could be used to process images or videos of unused and faulty products accumulated in the user's house; then, for example, products made of critical raw materials \cite{graedel2015criticality,CriticalityEU,CriticalityUS} could be collected in agreement with the householder for their recycling/re-manufacturing.                
      
Based on the observation that food items often have ingredients distributed in slices (i.e. layers), the authors of \cite{martinel2018wide} propose using ``slice'' convolutional kernels in a CNN \cite{lecun1999object} to improve the model classification accuracy. The authors also point out that such an accurate model to date requires memory and computational costs too high to be implemented on devices with limited resources (e.g. on mobile phones as in \cite{kawano2015foodcam}). The idea of adapting the neural layers to the type of target items can be seen as a method to embed a-priori knowledge in the layers architecture; an alternative approach could be adding a-priori knowledge through fine-grained classes able to catch minor differences between target items (e.g. ravioli vs. dumplings, mobile phone of brand A vs. brand B) by formulating a multi-task loss function as proposed in \cite{wu2016learning}.  

Transfer learning \cite{lu2020knowledge} has been shown to improve the food classification accuracy of neural models compared to models trained from scratch \cite{sun2019exploring,xiao2019deep}. Similarly, the features learned by a network trained on datasets with images of different types of objects could be fine-tuned (either the whole network or just part of it) on smaller datasets of images of products whose materials are of particular interest, e.g. critical raw materials \cite{graedel2015criticality,CriticalityEU,CriticalityUS}. To exploit transfer learning, over the years, large neural models have been developed, trained on large datasets, made publicly available and ready to use (see \revision{Subsubsection \ref{sub:DatasetsAndModels} for details}).    

Data augmentation \cite{shorten2019survey} could be applied to increase the number of images available for training. An example application for food recognition is proposed in \cite{lu2016food}, where new training images are generated rotating, translating and rescaling the original ones. Similarly, as in \cite{OurPaperIFAC2020}, data augmentation could be used with images of trash pieces to accurately classify their material composition (e.g. paper, plastic).

\subsubsection{From food portion to material volume estimation} 
Given the similarity between processing food and non-food items, below we discuss relevant papers on food calories or food volume estimation.     

\begin{figure}
\centering
\includegraphics[width=0.7\textwidth]{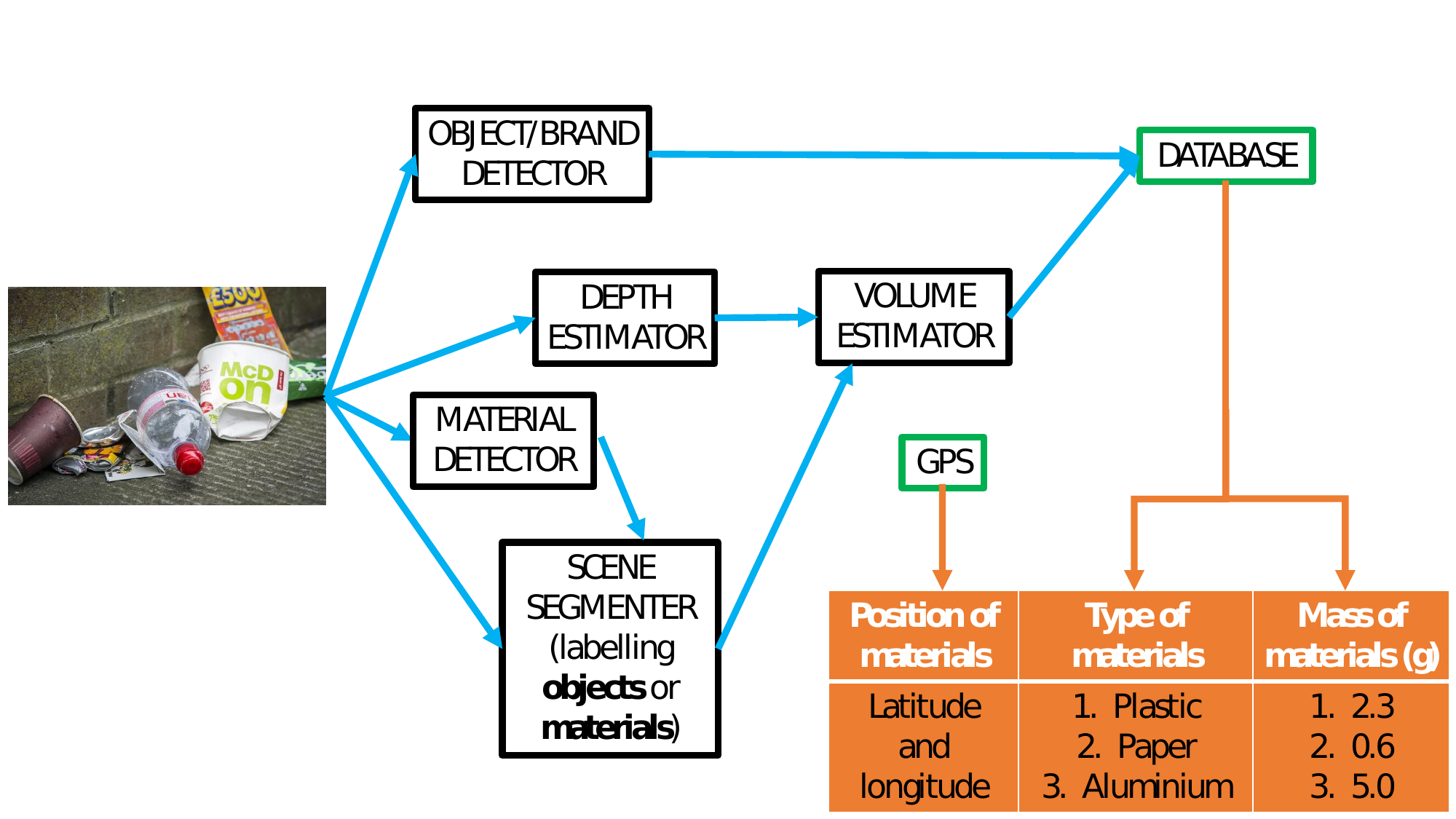}
\caption{An example of the system proposed in \cite{meyers2015im2calories} adapted to measuring material compositions instead of food calories.}
\label{fig:AdaptationFromGoogle_cropped}
\end{figure}

In \cite{meyers2015im2calories} the food volume is estimated requiring a single RGB image using the CNN architecture of \cite{eigen2015predicting} to work as a virtual depth camera; then, the depth map is converted into a voxel representation; semantic segmentation \cite{yu2018methods} is performed to identify the pixels corresponding to food and finally, in combination with the voxel representation, the volume of food is estimated. This approach, as pointed out by the developers, simply requires a single image collected ``from the wild'', hence it is particularly flexible; however, the whole system is quite complex as it combines multiple tasks, each one with its own complexity and with the accuracy/robustness of the whole system depending on the accuracy/robustness of all its components: open-world recognition (i.e. detecting food items from a generic scene), depth measurements from a single RGB image, 3d voxel representation from a 2d image, and scene segmentation. As pointed out by the authors, their system requires further development. Our interest in their system is its application to measure materials. An example of how the system could be adapted to our case is illustrated in Fig. \ref{fig:AdaptationFromGoogle_cropped}.           

In \cite{fang2018single} a generative adversarial network \cite{goodfellow2014generative} is used to map the input image depicting a food scene into the corresponding, pixel-by-pixel, energy content. The result is that their model reads the RGB food image and returns as output the energy (i.e. calories) at each pixel, e.g. a pixel without food has zero calories, a pixel belonging to broccoli corresponds to an energy weight lower than a pixel belonging to meat; hence, the energy weight can be seen as an energy density in $J/m^3$. In common with \cite{meyers2015im2calories}, this approach requires a single input image; a difference is in the fact that here the real size of the food is reconstructed using a marker of known size included in the food image rather than using a neural model as a virtual depth camera. Inspired by this work, an MMU could be implemented training a generative model to map the input RGB image of an object to the corresponding pixel-by-pixel density of material in $kg/m^3$; then, the summation of all the weight-densities overlapping each mask of the semantically segmented image gives an estimate of the mass of each labeled material.   

An approach based on stereo vision is proposed in \cite{dehais2016two}, hence it requires two images of the food item taken from two different views; after a test on the best compromise between accuracy and efficiency, SURF is chosen for feature extraction; the system requires a reference object placed next to the food to reconstruct the real sizes, and the food should be placed inside an elliptical, flat plate, i.e. bowls are not permitted. Compared to \cite{meyers2015im2calories}, this work has the advantage of being implementable on a computer with limited performance (e.g. a mobile phone). The simplest version implemented in \cite{meyers2015im2calories} requires the user to select the best labels and exploits the knowledge of the menu of the restaurant the dish belongs to, whereas their flexible and highly automated more complex version is, according to the authors, at a preliminary stage. Note that \cite{dehais2016two} focuses on food portion volume estimation without evaluating the calories/energy content. Its application could be adapted to estimate the volume of the components of a manufacturing product and then, knowing the density of detected materials, converted into masses. Such an MMU could run in a mobile phone. Bearing in mind the analogy between volume estimation of food and the volume estimation of manufacturing products to approximate their mass, the interested reader should refer to \cite{min2019survey} for further reading on food volume or calories estimation.

\subsubsection{From food recipes to material composition of products}
An output that MMUs should provide is the list of materials making a target product. In general, we see two approaches to making the system capable of providing such information: (1) embedding it inside the mathematical model during the supervised training phase so that the model learns it (e.g. segmenting the scene labeling the materials as in \cite{bell2015material}); (2) providing the MMU with access to a list of materials stored in a database (e.g. as in \cite{meyers2015im2calories}, where the recipe of a detected food item is retrieved from the restaurant menu). Both approaches require the MMU developer to know the material composition of the product of interest. Hence, below we list five methods to collect information about the recipes of manufacturing products, some of which are used in food computing \cite{min2019survey} to collect food recipes: 
\begin{itemize}
\item{cookbook-like, i.e. looking at books describing the manufacturing process and material composition of the target product such as the ones in Table \ref{Tab:Cookbooks};}
\item{websites, e.g. the manufacturer website providing the technical specifications of its products, websites collecting material compositions similar to the ones created for recipes (e.g. Allrecipes, RecipeSource);}
\item{research papers reporting the material composition of specific products such as the ones in Table \ref{Tab:CompositionPapers};}
\item{performing a chemical composition analysis of the product of interest;}
\item{documentaries such as \revision{``How It's Made'' \cite{HowItsMadeWikipedia}}.} 
\end{itemize}
\begin{table}
\caption{Examples of books covering the raw material composition of complex materials or products.\\}
\label{Tab:Cookbooks}
\centering
\begin{tabular}{c@{\hskip 0.3in}c@{\hskip 0.3in}c@{\hskip 0.3in}c}
Material or product & Book & Publication year & Publisher\\
\hline
\revision{C}arbon and graphite materials & \cite{frohs2021book} & 2021 & Wiley\\
\revision{R}ubber & \cite{dick2014raw} & 2014 & Hanser Publications\\
\revision{N}onwoven fabrics & \cite{albrecht2006nonwoven} & 2006 & Wiley\\ 
\revision{E}nergy systems & \cite{bleicher2020material} & 2020 & Elsevier\\
\revision{E}lectrical and electronic materials & \cite{gupta2015advanced} & 2015 & Wiley\\
\revision{P}rinted electronics & \cite{cui2016printed} & 2016 & Wiley\\ 
\revision{P}aper & \cite{bajpai2018biermann} & 2018 & Elsevier\\ 
\revision{P}lastics & \cite{brydson1999plastics} & 1999 & Elsevier\\
\revision{A}utomobile bodies & \cite{davies2012materials} & 2012 & Elsevier\\
\revision{M}iscellaneous & \cite{allwood2012sustainable,allwood2015sustainable} & 2012, 2015 & UIT Cambridge\\
\revision{F}ood packaging & \cite{piergiovanni2016food} & 2016 & Springer\\
\revision{B}uildings & \cite{duggal2017building} & 2017 & Routledge\\
\revision{A}ircrafts & \cite{mouritz2012introduction} & 2012 & Woodhead Publishing\\
\hline    
\end{tabular}
\end{table}
\begin{table}
\caption{Examples of published research articles reporting the results from an analysis of the material composition of manufactured products. The rare Earth elements, as defined by the British Geological Survey \cite{REEdefinition}, are highlighted in bold.\\}
\label{Tab:CompositionPapers}
\centering
\begin{threeparttable}
\begin{tabular}{ccc}
Analyzed product & Article & Reported chemical elements and materials\tnote{1}\\
\hline
LCD\tnote{2} & \cite{lahtela2019novel} & In, Sn, Al, Fe, Cu, Zn\\ 
\revision{F}luorescent lamp & \cite{rabah2008recyclables} & Ca, \textbf{Y}, \textbf{Eu}, glass, Al, resin, Cu, Ni, brass, Hg, W, others\\
\revision{C}omputer monitor & \cite{resende2010study} & \textbf{Y}, In, \textbf{Ce}, \textbf{Eu}, Al, Si, S, K, Ca, Mn, Fe, Zn, Sr, Zr, Ir, Pd, Ba, Pb\\
\revision{M}obile phone & \cite{kim2018metal} & Fe, Cu, Zn, Ni, \textbf{Nd}, \textbf{Pr}, Cr, Sn\\
\revision{H}eadset, HDD, SSD\tnote{3} & \cite{thiebaud2018our} & \textbf{Nd}\\
\revision{V}ehicle battery & \cite{fishman2018implications} & Li, Al, Cr, Mn, Fe, Co, Ni, Cu, \textbf{La}, \textbf{Ce}, \textbf{Pr}, \textbf{Nd}, \textbf{Gd}, \textbf{Tb}, \textbf{Dy}, \textbf{Er}\\
PCB\tnote{4} & \cite{xiu2010materials} & Al, Cu, Zn, Fe, Mn, Sn, Pb, Ag, Be, Ti, Sb, As, Ni, Mg, Ba, Cd\\
HDD & \cite{icsildar2018electronic} & \textbf{Dy}, \textbf{Nd}, \textbf{Pr}\\
LCD & \cite{icsildar2018electronic} & In, \textbf{Y}\\
\revision{C}ordless phone & \cite{hageluken2006improving} & Cu, Al, Fe, plastics, Ni, Pb, Sn, Ag, Au, Pd, others\\
DVD player\tnote{5} & \cite{hageluken2006improving} & Cu, Al, Fe, plastics, Ni, Pb, Sn, Ag, Au, Pd, others\\
\revision{C}alculator & \cite{hageluken2006improving} & Cu, Al, Fe, plastics, Ni, Pb, Sn, Ag, Au, Pd, glass, others\\
\revision{P}rinter PCB & \cite{yoo2009enrichment} & Cu, Al, Ni, Fe, Sn, Pb, Zn, Co, Ti, Ag, Au\\
\revision{L}aptop PCB & \cite{icsildar2016two} & Cu, Fe, Al, Ni, Zn, Pb, Cr, Au\\
\hline    
\end{tabular}
\begin{tablenotes}\footnotesize
\item[1] Symbols of the chemical elements taken from the periodic table
\item[2] LCD: liquid-crystal display
\item[3] HDD: hard-disk drive; SSD: solid-state drive
\item[4] PCB: printed circuit board
\item[5] DVD: digital video disk
\end{tablenotes}
\end{threeparttable}
\end{table}      
Considering that the recipe of a food item is a list reporting the types and masses of materials/ingredients that compose the desired food type, research in food ingredient recognition could be useful. For example, \cite{chen2016deep} is one of the first works focusing on ingredient recognition rather than food category recognition, which was the main research trend at that time. The authors proposed a multi-task learning technique to recognize, at the same time, both the food type and its ingredients because the features learned to solve one task improve the recognition accuracy in the other task and vice versa and therefore it improves the robustness of the whole system. The authors also pointed out and investigated two key design issues which are valid for MMUs as well: first, ``given a deep neural network architecture to be trained in a multi-task fashion, to what extent should the two tasks share network layers?''; and second, ``should the tasks be solved in series, i.e. the output scores of a task are the input to the other one, or in parallel?''. To answer these questions, four different neural architectures were derived by modifying the VGG 16-layer network \cite{simonyan2014very}. Along with the multi-task learning, a region-wise ingredient recognition approach is proposed in \cite{chen2020study} which consists of dividing the input image into many small regions and performing ingredient recognition on each single region. Hence, with region-wise learning, the system also localizes the ingredient spatial distribution over the image along with the list of ingredients. Further works on ingredient recognition can be found in \cite{min2019survey}.\\
\fbox{\begin{minipage}{33.9em}
\revision{\textbf{Subsection take-aways:} computer vision on food recognition, food portion estimation and ingredient recognition can be adapted to focus on non-food products; since $mass$ = $density$ $\times$ $volume$, volume estimation is related to mass estimation; both hand-crafted-feature-based and CNN-based methods have been proposed in food computing; a challenge in volume estimation is the accurate extraction of 3d information from a single 2d image; while stereo vision requires two input images, a main challenge is requiring just a single image without reducing accuracy.}
\end{minipage}}

\subsection{Towards a Sensor Network for Material Stock Monitoring}\label{sub:SensorNetwork}
\noindent
\fbox{\begin{minipage}{33.9em}
\textbf{Definition 2.} \emph{A manufacturing network is a set of locations/buildings connected by the exchange of material, e.g. raw material reservoirs, manufacturers, shops, houses, waste sorting centers, recycling centers, landfills.}
\end{minipage}} 
\vspace{0.05in}

An \emph{analogy} between water networks and manufacturing networks can be seen considering both an intuitive and a physical explanation. The intuitive explanation is that both networks are made of nodes (i.e. compartments) that exchange materials (e.g. water, aluminum, plastic) over time and space; the physical explanation is based on the force-voltage physical analogy, also known as Maxwell's analogy \cite{borutzky2009bond}, as detailed in Table \ref{Tab:Analogy}. Note that the two networks have a different \emph{effort variable} because a fluid flows within a water network, whereas materials (including products, which are combinations of materials) ``flow'' within a manufacturing network. Another difference is that a hydraulic network confines the fluid in pipes, whereas a manufacturing network moves the materials using transport systems, e.g. trucks, airplanes, ships.        
\begin{table}
\caption{Analogy between water and manufacturing networks; the SI units are between squared brackets.\\}
\label{Tab:Analogy}
\centering
\begin{tabular}{c|c@{\hskip 0.1in}c@{\hskip 0.1in}c}
& Displacement & Flow variable & Effort variable\\
\hline
\makecell{Water \\ network} & \makecell{\revision{W}ater \\ mass [$kg$]} & \makecell{\revision{M}ass flow \\ rate [$kg/s$]} & \revision{P}ressure [$N/m^2$]\\ 
\hline
\makecell{Manufacturing \\ network} & \makecell{\revision{M}aterial \\ mass [$kg$]} & \makecell{\revision{M}ass flow \\ rate [$kg/s$]} & \revision{F}orce [$N$]\\
\hline 
\end{tabular}
\end{table}

\begin{figure*}[t]
\centering
\includegraphics[width=0.8\textwidth]{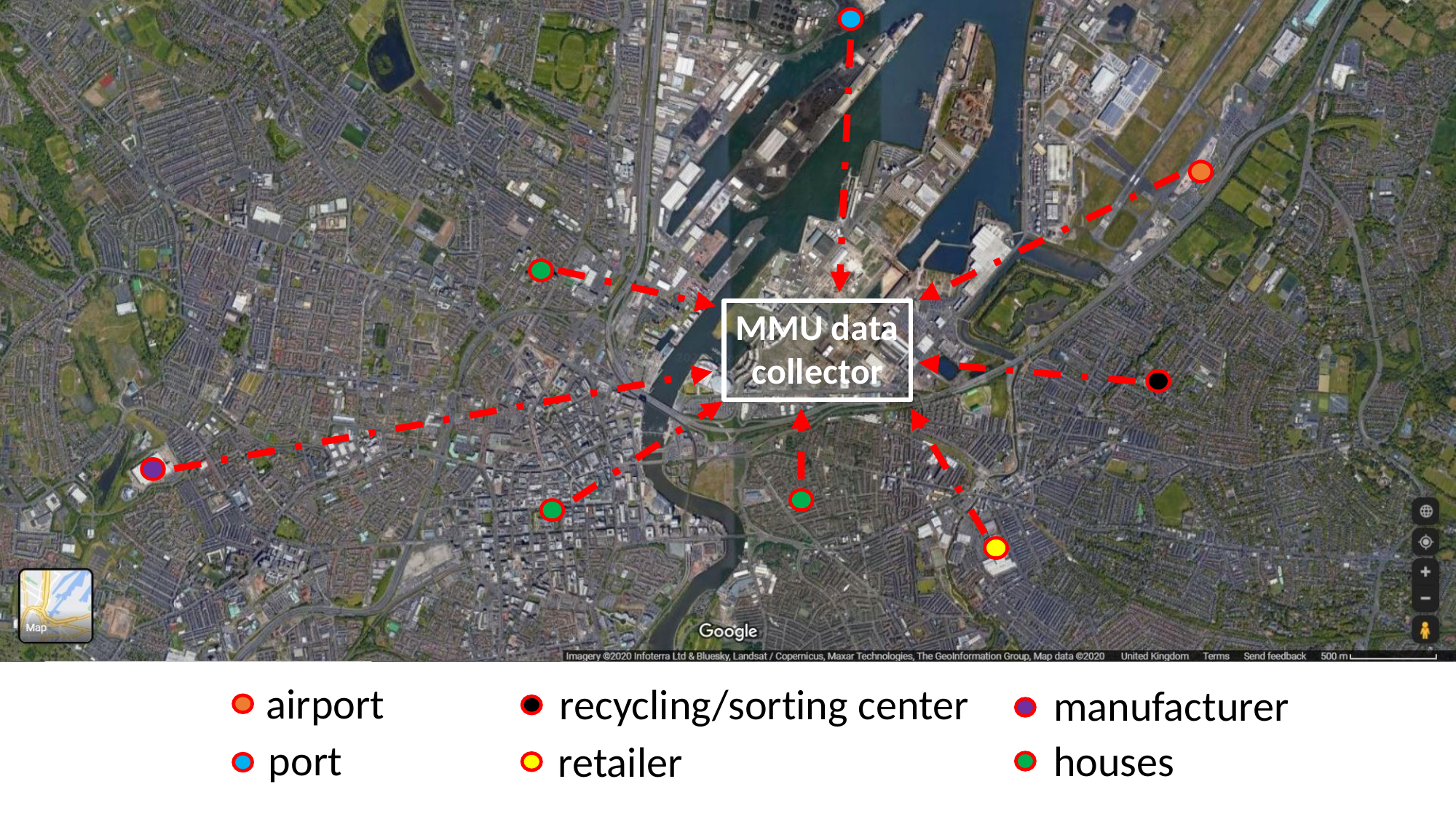}
\caption{\revision{Example of the MMU sensor network introduced in Fig. \ref{fig:OverviewScheme} that could be implemented in the city of Belfast, Northern Ireland: material measurements from multiple units are collected in a central database referred to as ``MMU data collector'' in this figure, which can extend current material stocks and flows databases such as the YSTAFDB \cite{myers2019ystafdb} for MFA studies on the implementation of circular material flows such as \cite{mehta2022using,millette2019materials,MFA-Europe}; a GPS could provide the location of the measured materials at different points as introduced in Fig. \ref{fig:AdaptationFromGoogle_cropped}.}}
\label{fig:MMUnetwork}
\end{figure*}

The physical analogy between water and manufacturing networks suggests that MMUs could be used as a sensor network to monitor the flow of materials as the system proposed in \cite{stoianov2008sensor} for water networks. Multiple platforms equipped with MMUs could form a sensor network for material stock monitoring. An example of the system is shown in Fig. \ref{fig:MMUnetwork} considering the design of an MMU sensor network for a city using Belfast as an example.
Below we list four types of platform upon which an MMU could be implemented to realize an MMU sensor network.

\subsubsection{MMU in trucks and bins} % focus on CV + bins + trucks
In \cite{hannan2011radio,arebey2012solid,hannan2013automated} a computer vision system is proposed to process internal images of bins. In particular, in these works: first, the camera is installed on the truck (i.e. a smart truck) and used by a worker when the truck approaches the bin; second, the image processing returns information only about the waste level, not the waste type and mass as an MMU would do. A smart truck is also proposed in \cite{rad2017computer,zhang2019urban} for urban street waste mapping and cleanliness assessment (refer to \cite{hannan2015review} for a review of the information and communication technologies used for waste management). 

The use of multiple sensors for smart bins has been considered over the years and the interested reader is referred to the review paper by Soni et al. \cite{soni2017smart}, which organizes and details the state-of-the-art in 2017. This review also proposes a framework for a waste management system based on smart bins leveraging the cloud and wireless communication to create a network of smart bins. Therefore, two aspects are particularly relevant in the context of our paper: first, Soni et al.'s network system based on smart bins is a sub-network of the system we propose here for material stock monitoring; and second, integrating into smart bins the computer-vision-enabled material measurements provided by MMUs could improve the accuracy of a smart bin network.

\subsubsection{MMU in autonomous sorting systems} % focus on CV + sorting plants
A vision-based sorting system processes the types of material for which it has been trained to detect and using a larger and richer training set, in general, improves the adaptability of the system to work with other materials. In fact, if a new class of material needs to be detected, samples belonging to that class must be learned by the autonomous sorter. In contrast, techniques based on mechanical and magnetic principles can sort only a specific type of material according to its physical/chemical properties (e.g. magnetic drums, air separators, triboelectrostatic separators) \cite{gundupalli2017review}. Examples of works integrating computer vision in autonomous sorting systems have been proposed in \cite{lukka2014zenrobotics} for demolition waste, in \cite{laszlo2019sorting} for electronic waste, in \cite{blasco2009development} for pomegranate arils, in \cite{pervsak2020vision} for plastic granulate, in \cite{seredkin2019development} for municipal solid waste, in \cite{tessier2007machine} for rock mixture composition and in \cite{kumar2018material} in a patent application. The low use of advanced computer vision based on deep learning suggests that a significant improvement of autonomous sorting systems is possible.

\subsubsection{MMU in mobile phones} 
As far as we know, the only work proposing a mobile phone application for waste detection is \cite{mittal2016spotgarbage}: the pre-trained model AlexNet \cite{krizhevsky2012imagenet} is fine-tuned on a garbage-focused dataset using a GPU and then simplified to be implemented in smartphones; after an analysis of the image processing time required by the different components of the system, the model with the best compromise between classification accuracy, image processing time and memory size is chosen.  

As discussed in \revision{Subsection \ref{sub:OnFood}}, computer vision systems proposed for food classification and calories estimation are related to material classification and mass estimation, respectively. In \cite{meyers2015im2calories} the system prototype runs on mobile phones and requires the user to provide a single image, whereas \cite{gao2019musefood,dehais2016two} need two images. An interactive application is proposed in \cite{kawano2015foodcam}, whereas the authors of \cite{pouladzadeh2017mobile} delegate the most complex tasks such as food recognition to a cloud server instead of to the phone CPU.

Implementing MMUs on mobile phones is particularly challenging because the computationally expensive task of estimating the mass from images is necessary, whereas it may be avoided in automated sorting facilities or bins through the use of weight scales; moreover, mobile phones have limited computing performance compared to the hardware that can be used in sorting facilities, e.g. GPUs. On the other hand, a mobile phone has the advantage of being portable and cheap. Potentially any owner of a mobile phone could collect material measurements through an MMU mobile application or send images/videos to a central server running an MMU.\\
\fbox{\begin{minipage}{33.9em}
\revision{\textbf{Subsection take-aways:} MMUs can be implemented on different platforms to map and quantify the material stocks in a target area; the GPS provides the material location; to date vision systems are widely used for robotic applications (e.g. inspection, sorting) and not for resources mapping; implementing vision systems on a mobile phone poses constraints on the computing performance, but has the advantages of portability and affordability.}
\end{minipage}}

\subsection{Hands-On Deep Learning for MMU Development}\label{sec:HandsOn}
\subsubsection{\revision{Getting started}}
\revision{The first step to using or developing deep learning algorithms is selecting an appropriate programming language. Currently there are two main choices: MATLAB and Python. MATLAB has a toolbox dedicated to deep learning named Deep Learning Toolbox \cite{MAT-DLtoolbox} which provides two development environments: a script-based one and the Deep Network Designer. The former is pure coding, therefore is more flexible, but less user-friendly. In contrast, the latter is particularly easy to work with as it has an intuitive user interface. The use of the Deep Learning Toolbox requires a license. Python libraries, in contrast, are open source. The main Python libraries for deep learning are PyTorch \cite{stevens2020deep} and TensorFlow/Keras \cite{TFtutorial}. Python users can run on a free GPU service through Google Colaboratory (aka ``Colab''). Usually, the latest algorithms are developed in Python and subsequently integrated by MathWorks into its toolbox; the advantage of using the latest algorithm is usually at the cost of a less documented and less user-friendly code.}
 
\subsubsection{Public datasets, pre-trained models and transfer learning}\label{sub:DatasetsAndModels} 
In general, the model of an MMU can be defined in two ways: using a pre-trained model or training a model from scratch. Usually the model reaches a good accuracy if it is at least fine-tuned with images depicting the domain of application. Hence, in Table \ref{Tab:Datasets} we list the source paper and the download website of several publicly available datasets containing images of manufacturing products or waste items that could be useful to train/fine-tune MMU models.
\begin{table}
\caption{Publicly available datasets with images of materials and manufacturing products in use or as waste.\\}
\label{Tab:Datasets}
\centering
\begin{threeparttable}
\begin{tabular}{c@{\hskip 0.001in}c@{\hskip 0.001in}c@{\hskip 0.001in}c}
Dataset name & Reference & Original task & Waste focused?\\
\hline
Caltech 101\tnote{1} & \cite{fei2006one} & Object recognition & No\\ 
Caltech 256\tnote{2} & \cite{griffin2007caltech} & Object recognition & No\\  
COIL100\tnote{3} & \cite{nene1996object} & Object recognition & No\\
COCO\tnote{4} & \cite{lin2014microsoft} & Object recognition & No\\
MINC\tnote{5} & \cite{bell2015material} & Material recognition & No\\
ADE20K\tnote{6} & \cite{zhou2017scene} & Object recognition & No\\
Open Images\tnote{7} & \cite{OpenImages2} & Object recognition & No\\
trashnet\tnote{8} & \cite{yang2016classification} & Material recognition & Yes\\
TACO\tnote{9} & \cite{taco2020} & Material/object recognition & Yes\\
MJU-Waste\tnote{10} & \cite{wang2020multi} & Object recognition & Yes\\
Flickr Material Database\tnote{11} & \cite{sharan2009material} & Material recognition & N/A\\ 
GINI\tnote{12} & \cite{mittal2016spotgarbage} & Object recognition & Yes\\
CUReT\tnote{13} & \cite{dana1999reflectance} & Material recognition & N/A\\
ImageNet\tnote{14} & \cite{deng2009imagenet} & Material/object recognition & N/A\\
DTD\tnote{15} & \cite{cimpoi2014describing} & Texture recognition & No\\
recybot\tnote{16} & \cite{jahanian2019see} & Object recognition & Yes\\  
\hline    
\end{tabular}
\begin{tablenotes}\footnotesize
\item[1] \url{http://www.vision.caltech.edu/Image_Datasets/Caltech101/#Description}
\item[2] \url{http://www.vision.caltech.edu/Image_Datasets/Caltech256/}
\item[3] \url{https://www1.cs.columbia.edu/CAVE/software/softlib/coil-100.php}
\item[4] \url{https://cocodataset.org/#home}
\item[5] \url{http://opensurfaces.cs.cornell.edu/publications/minc/}
\item[6] \url{https://groups.csail.mit.edu/vision/datasets/ADE20K/}
\item[7] \url{https://storage.googleapis.com/openimages/web/index.html}
\item[8] \url{https://github.com/garythung/trashnet}
\item[9] \url{http://tacodataset.org/}
\item[10] \url{https://github.com/realwecan/mju-waste}
\item[11] \url{http://people.csail.mit.edu/celiu/CVPR2010/FMD/index.html}
\item[12] \url{https://github.com/spotgarbage/spotgarbage-GINI/blob/master/README.md}
\item[13] \url{https://www1.cs.columbia.edu/CAVE/software/curet/index.php}
\item[14] \url{http://www.image-net.org/}
\item[15] \url{https://www.robots.ox.ac.uk/~vgg/data/dtd/index.html}
\item[16] \url{https://github.com/MIT-MRL/recybot}
\end{tablenotes}
\end{threeparttable}
\end{table}
If the choice is to use pre-trained models rather than training from scratch, the links to pre-trained models available in some machine learning libraries are: \revision{MATLAB \cite{MAT-pretrainedModels}, PyTorch \cite{PyTorch-pretrainedModels} and TensorFlow \cite{TF-pretrainedModels}.} Guidance on how to exploit a pre-trained model can be found in the area of research known as \emph{transfer learning} \cite{lu2020knowledge}, which typically deals with transferring into a new model the machine knowledge/experience contained in another model previously trained on a dataset different from the one of interest.

\textbf{Transfer learning.} In transfer learning, the task to be solved with the new model (e.g. recognizing materials) is referred to as \emph{target task/domain}, whereas the task solved with the previously trained model and from which the knowledge has to be transferred is referred to as \emph{source task/domain} (e.g. recognizing common objects). There are two typical situations in which transfer learning is particularly useful. One situation is when the developer of the vision system has no or very few images of the target domain, which occurs when the task is very specific (e.g. recognizing laptops of a particular brand); hence, training a neural model from scratch with just those samples yields an inaccurate model. The other situation is when one needs a large neural model to solve a complex target task, but due to time constraints its training is not practically feasible \cite{lu2020knowledge}. In general, the following key considerations should be kept in mind about transfer learning:
\begin{itemize}
\item{the best accuracy possible is achievable by training from scratch a model using many samples of the target domain, which means not performing any knowledge transfer;} 
\item{the knowledge transfer effectiveness is higher when the target and the source domains are strongly related and decreases with the reduction of the source-target similarity (e.g. images of cats are more similar to images of dogs rather than images of plastic bags);}
\item{when transferring features from the layers of a pre-trained CNN, the features from the first layers contain general visual traits and therefore are more transferable to different domains, whereas the deepest layers are more optimized for the source task \cite{yosinski2014transferable};}
\item{transferring knowledge from sources unrelated to the target task may cause the negative transfer effect \cite{rosenstein2005transfer}.}
\end{itemize}

\subsubsection{Lack of training images and class imbalance}\label{subsec:ClassImb}
When designing a vision system, one may experience a lack of images or the class imbalance problem, which consists of having some classes with a large number of samples and other classes with few samples. Both the lack of training samples and class imbalance result in inaccurate models \cite{shorten2019survey}. Some transfer learning techniques have been proposed to create a larger training dataset exploiting different source domains \cite{lu2020knowledge} and these techniques could also be adapted to address class imbalances considering only specific target classes.

\textbf{Data augmentation.} \emph{Data augmentation} techniques are used in computer vision specifically to extend the training dataset size \cite{shorten2019survey}. The key idea behind data augmentation is to generate new images that are variants of the original training samples by adding to them particular effects such as rotations, translations, scale variations, lighting variations, occlusions. Thus, even though the initial set of images depicts only a few scenes, the set is enriched by adding properly altered samples. Along with traditional hand-crafted augmentation techniques that generate the aforementioned types of effects, recent advances in deep learning led to the development of further methods. In particular, the ones based on generative adversarial networks (GANs) were shown to be particularly effective for their computational efficiency and quality of results \cite{shorten2019survey} (see \cite{wang2021generative} for a recent review on GANs for computer vision). Since GAN-generated samples are not as predictable as the ones generated using traditional methods (e.g. rotations), an effective way to extend the size of the training set is to use both traditional and non-traditional approaches and then collect together their outputs. To assess the quality of the synthetic samples, a visual Turing test was performed in \cite{frid2018gan} with medical images by asking two expert radiologists to distinguish between original and altered samples. The two radiologists labeled the GAN-generated images with 62.5\% and 58.6\% accuracy, respectively. Augmenting the training dataset requires that extra storage space is available and this aspect is particularly critical when very large datasets are enlarged (e.g. millions of images). Alternatively, one may implement an on-line data augmentation, which generates the new samples during the model training. While this second strategy can save memory, it results in a longer training time \cite{shorten2019survey}.

\subsubsection{Computing platforms} 
Deep learning techniques are permitting the design of machines more accurate than ever for solving tasks such as classification of images or regression, which are tasks relevant for MMUs. However, deep learning machines are also more computationally demanding than traditional machine learning. Since high performance computers are not portable, an active area of research is investigating the most efficient strategies to exploit large neural models on portable, low performance devices such as sensored microcontrollers and smartphones \cite{chen2019deep}. 
The two opposite implementation paradigms are \emph{cloud computing} and \emph{edge computing} (optimal solutions may be a compromise between them depending on the specific situation).
Cloud computing consists of performing the computation on large and high performance computers that exchange data with smaller devices through the Internet. To have an idea of the size of such high performance machines, a Microsoft data center is 11.5 times the size of a football field \cite{hwang2017cloud}. With this approach, the portable device mainly transfers the measurements from the acquisition site to the cloud and reads the reply once the computation is completed. In contrast, edge computing seeks to perform all the required computations where the measurements are acquired, i.e. on the portable device. If the choice is to implement the MMU using cloud computing, popular platforms providing cloud services are: \revision{Amazon Web Services (AWS) \cite{AmazonWS}, Microsoft Azure \cite{MicrosoftAzure}, Google Cloud \cite{GoogleCloud} and IBM Cloud Services \cite{IBMcloud}.}
While cloud computing has the advantage of providing the maximum computing resources and, therefore, speed of execution, it has three main disadvantages which motivate the decentralization of the workload from cloud servers to edge devices: latency, scalability and privacy.
\begin{enumerate}
\item{Latency: moving the data from the edge device to the cloud is a time consuming activity which may compromise the system performance in real-time applications; for example, sending and processing a camera frame for a computer vision task could take more than 200 ms end-to-end on Amazon Web Services \cite{satyanarayanan2017emergence}.}
\item{Scalability: cloud platforms provide many servers to satisfy the computing demand of their users, but a large number of connected devices may overload the network and the data center resulting in issues such as communication delays.}
\item{Privacy: the exchange of data between different devices could compromise privacy if sensitive information is improperly used or intercepted.}
\end{enumerate}
If the choice is to develop MMU systems minimizing the use of the cloud, the following hardware and software are particularly suitable.
\begin{itemize}
\item{TensorFlow Lite is an open source machine learning library developed by Google optimized for smartphones and microcontrollers.}
\item{PyTorch Mobile is another open source machine learning library for edge computing and is developed by Facebook.}
\item{The Nvidia Jetson production line are development kits that integrate GPUs on a microcontroller and have primarily been designed to speed up neural network algorithms on portable devices.}
\item{Two microcontrollers known for their ease of use are Raspberry Pi and Arduino; in particular, the development team of the latter is currently working to integrate machine learning functionalities and the Arduino Nano 33 BLE Sense is the recommended board to get started.}
\end{itemize} 
\fbox{\begin{minipage}{33.9em}
\revision{\textbf{Subsection take-aways:} TensorFlow/Keras, PyTorch and the MATLAB Deep Learning toolbox are the main libraries for feature-learning-based (i.e. CNN-based) computer vision; relevant public datasets are summarized in Table \ref{Tab:Datasets}; the links to public pre-trained models in MATLAB, PyTorch and TensorFlow/Keras are provided; a GPU speeds-up CNN-based algorithms compared to a CPU (both training and inference); for Python users, Google Colaboratory may be a valid option to access a ready-to-use GPU; data augmentation techniques can extend the training dataset size; the Nvidia Jetson production line provides GPUs on microcontrollers.}
\end{minipage}}

\section{\revision{Discussion}}\label{sec:Challenges}
\subsection{Challenges}
The design of a material stock monitoring system that accurately and in real-time gives the types, masses and positions of materials in a target area involves multiple challenges. In general, our recommendation is to begin working on single devices and then connect multiple units only when the material measurements are of satisfactory accuracy. This size-increasing rationale is indeed followed in writing \revision{Subsections \ref{sub:OnMaterials}, \ref{sub:OnFood} and \ref{sub:SensorNetwork} with that order}: from material recognition, which is Component 1 of an MMU unit, we then add the object recognition, the volume estimation and, at the very end, the idea of networked units that send material stock information to a central data collector. A two-task learning process for both object and material recognition can be guided by previous works on food computing \cite{chen2016deep,chen2020study}, whereas how to design a three-task learning process that involves also mass estimation is an open question. A first prototype of such a multi-component vision system is proposed in \cite{meyers2015im2calories} for food computing, but, as pointed out by the authors, there are many challenges for computer vision researchers especially when it comes to implementing the system on small devices such as smartphones. Some examples of these challenges are recognition of the material considering small regions of the image (i.e. fine-grained recognition), recognition considering images taken in the open world rather than in a laboratory setting, depth estimation from a single RGB image and minimization of latency for real-time applications. The challenges for computing accurate material measurements are in computer vision, which is currently a very active and advancing research topic as it is one of the main areas of application of deep learning. At the same time, edge computing research \cite{chen2019deep} should proceed to permit the implementation of high performance vision systems on small devices. In this way, an MMU could easily run on a smart bin, a smart truck and a smartphone. Once the vision algorithm of single units is able to provide reliable material measurements, the main challenges will be in the network communication system to transfer the measurements in real-time from different nodes.     

Along with the size-increasing rationale described above, another way to begin is by prioritizing critical materials, where the precise definition of ``critical'' may depend on the region or country of interest. For example, municipal electronic waste or electronic devices kept in private houses and no longer in use (e.g. due to being faulty or out of fashion), are a valuable source of rare Earth elements. Hence, first prototypes of MMU may focus on electrical and electronic items to improve the management of these critical non-renewable resources. \revision{Whether it is for electrical and electronic items or other materials such as waste plastics, the use of networked MMUs could make an important contribution in the transition to a circular economy, a barrier to which is ineffective collection and sorting of wastes \cite{mehta2022using}. Previous research by the authors has highlighted concerns over the ability of the public to properly sort waste streams \cite{mehta2021exploring}, as well as the need for better flow of data between manufacturers and waste collectors to reduce costs \cite{mehta2022using}. Networked MMUs could improve understanding of both waste sorting and overall material flows, which if implemented as part of a suite of measures, could support the changes needed for a circular economy.}

\textbf{Importance of benchmarking.} Regardless of the preferred research direction, benchmarking the resulting models in a standardized way permits their performance to be effectively assessed; benchmarking is standard practice among developers of classification systems, which makes it possible to rank models based on chosen \revision{metrics \cite{Benchmarking1,Benchmarking2}.} To advance the development of MMUs, examples of benchmarking metrics are: 
\begin{itemize}
\item{the model accuracy in \emph{waste} item classification}
\item{the model accuracy in \emph{material} classification}
\item{the model accuracy in \emph{volume} or \emph{mass} estimation} 
\item{the model computational complexity (e.g. seconds needed to process an image)}
\item{the model memory storage requirements (e.g. its size in MB).}
\end{itemize}

\subsection{\revision{Future Research and Development}} 
\begin{figure}
\begin{minipage}{\textwidth}
\centering
\includegraphics[width=\textwidth]{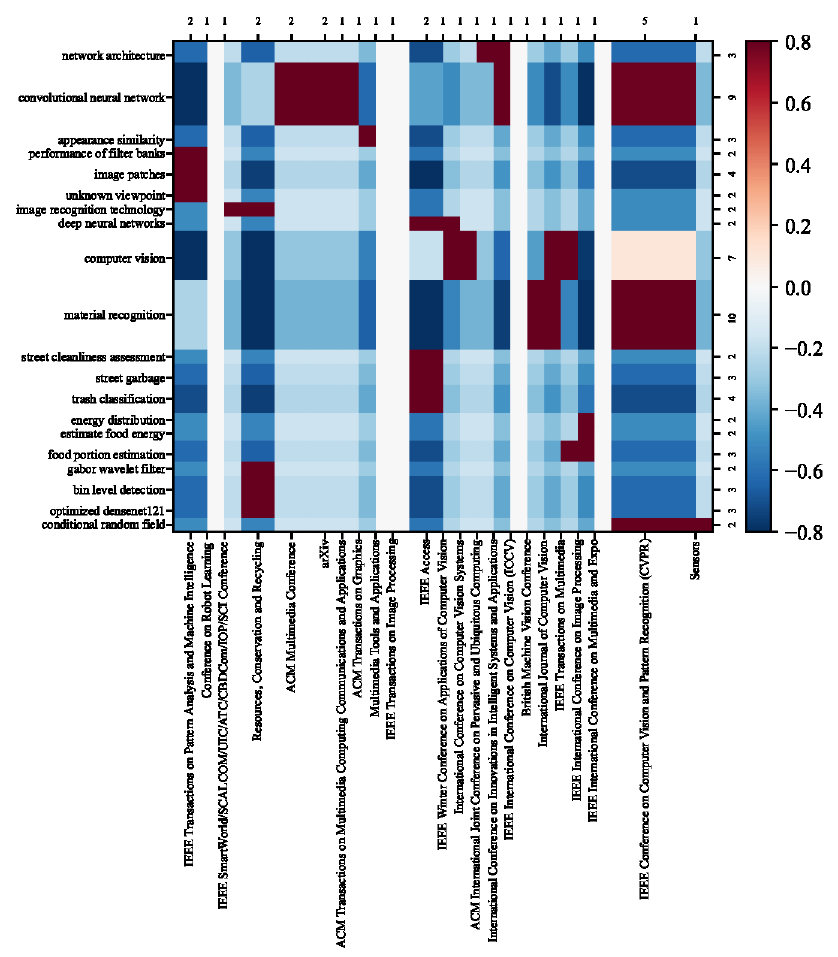}
\caption[...]{\revision{Contingency matrix of the papers listed in Table \ref{Tab:Summary}. Journal/conference names are on the x-axis and keywords extracted from the paper abstracts are on the y-axis. The numbers along the opposite axes indicate the frequency of each term. The rectangle sizes result from the frequency values while the rectangle colors define the correlation between each x-y pair, e.g. a dark red rectangle means that the abstracts of the papers published at the journal/conference along the x-axis often mention the keywords along the y-axis (``often'' is with respect to the average value). In contrast, a dark blue rectangle means rare mentions with respect to the average}\textsuperscript{a}.}\tiny\textsuperscript{a}We are most grateful to CorTexT platform (www.cortext.net) for its contribution in achieving the here-presented work powered by the use of the CorTexT Manager Application
\label{fig:ContingencyMatrix}
\end{minipage}
\end{figure}

Five main future research \revision{and development} paths are identified and summarized below: the first path essentially consists of implementing the most advanced recognition systems on platforms such as bins, mobile phones or sorting centers; the second, third and fourth paths are concerned with improving the three recognition systems already implemented on specific platforms; the last path consists of adapting systems from one platform to another (e.g. from mobile phones to smart trucks). \revision{Note that while developing an MMU mobile app is of practical use, doing so also addresses several fundamental challenges in computer vision research as pointed out by Myers et al. in the closing section of their ICCV paper \cite{meyers2015im2calories}: ``[…] it requires solving various problems, such as: fine-grained recognition […]; hierarchical label space […]; open-world recognition […]; visual attribute recognition […]; instance segmentation; instance counting; amodal completion of occluded shapes […]; depth estimation from a single image; information fusion from multiple images in real-time, on-device […]''.}  
\begin{enumerate}
\item{The systems developed in \cite{schwartz2019recognizing} and \cite{sharan2013recognizing} for material recognition are based on CNN and hand-crafted features, respectively, and are not implemented on specific platforms. Therefore, their implementation on mobile phones, smart bins and sorting centers could be investigated. Successively, the material recognizer could be combined with an object classifier to improve the system accuracy as done in \cite{hu2011toward} or in \cite{chen2016deep,chen2020study} with food items leveraging the analogy between ingredients (i.e. ingredient recognition) and material composition of non-food items (i.e. material recognition). Two preliminary questions arise with respect to the data-intensive CNN-based approach: ``Do I train the network from scratch?'' and ``Are the publicly available datasets sufficiently rich for the target application?''. If the answer to the second question is negative, a valuable contribution to the field would be the development and publication of a new dataset for this purpose as done in \cite{jahanian2019see}. Alternatively, the lack of training images could be mitigated by collaborating with a company that has collected them as done in \cite{ibrahim2019contaminet}.} 
\item{The mobile phone application of \cite{meyers2015im2calories} for food calories estimation is, according to the authors, at a preliminary stage. Their system is highly automated, but complex as it involves RGB map estimation, 3d voxel representation, open-world recognition and scene segmentation. Transferring the target application from food calories to material stock monitoring will result in a promising MMU.}
\item{While the system mentioned in the previous point is based on CNNs, the approach in \cite{dehais2016two} is based on hand-crafted features, therefore less computationally demanding. In general, \cite{meyers2015im2calories} could be seen as a more challenging research path to be ready for deployment later than the approach of \cite{dehais2016two}; however, the latter appears less promising in terms of both accuracy and flexibility.}
\item{The mobile phone application of \cite{mittal2016spotgarbage} could be improved, for example, by using a more advanced neural architecture with a similar computational complexity. Moreover, a high performance central server could communicate with the phone performing the most demanding tasks, i.e. exploiting cloud computing instead of edge computing.}
\item{The systems mentioned in the previous three points consider mobile phones. Their implementation in smart trucks could be investigated.} 
\end{enumerate}  
To help the reader with developing the five lines of research or identifying different ones, Table \ref{Tab:Summary} summarizes selected works covered in this review.
\begin{table}
\caption{Summary of selected works covered in this review. An empty entry in ``Platform'' means that the system was not deployed in a particular platform such as smartphones, bins, trucks or robots. ``Components'' indicates for which MMU component the work is most relevant, with the MMU components defined as in \revision{Subsection \ref{sec:MMUdefinition}}.\\}
\label{Tab:Summary}
\centering
\begin{tabular}{c@{\hskip 0.1in}c@{\hskip 0.1in}c@{\hskip 0.1in}c@{\hskip 0.1in}c}
Work & Approach & Application & Platform & Components\\
\hline
\revision{\cite{standley2017image2mass}} & \revision{Feature learning} & \revision{Mass estimation} & \revision{-} & \revision{3}\\
\cite{schwartz2019recognizing} & Feature learning & Material recognition & - & 1\\
\cite{sharan2013recognizing} & Hand-crafted features & Material recognition & - & 1\\
\cite{wang2020multi} & Feature learning & Waste detection & - & 1, 2\\
\cite{hu2011toward} & Hand-crafted features & Material recognition & - & 1\\
\cite{liu2010exploring} & Hand-crafted features & Material recognition & - & 1\\
\cite{varma2008statistical} & Hand-crafted features & Material recognition & - & 1\\
\cite{bircanouglu2018recyclenet} & Feature learning & Material recognition & - & 1\\
\cite{ibrahim2019contaminet} & Feature learning & Contamination detection & - & 1, 2\\
\cite{lagunas2019similarity} & Feature learning & Material similarity & - & 1\\
\cite{bell2015material} & Feature learning & Material recognition & - & 1\\
\cite{taco2020} & Feature learning & Waste recognition & - & 1, 2\\
\cite{cimpoi2014describing} & Hand-crafted features & Texture recognition & - & 1, 2\\
\cite{wu2009fast} & Hand-crafted features & Food calories estimation & - & 2, 3\\
\cite{yang2010food} & Hand-crafted features & Food recognition & - & 2\\
\cite{kawano2015foodcam} & Hand-crafted features & Food recognition & Smartphone & 2\\
\cite{martinel2018wide} & Feature learning & Food recognition & - & 2\\
\cite{wu2016learning} & Feature learning & Food recognition & - & 2\\
\cite{meyers2015im2calories} & Feature learning & Food calories estimation & Smartphone & 2, 3\\
\cite{fang2018single} & Feature learning & Food calories estimation & - & 3\\
\cite{dehais2016two} & Hand-crafted features & Food volume estimation & - & 3\\
\cite{chen2016deep} & Feature learning & Ingredient and food recognition & - & 1, 2\\
\cite{chen2020study} & Feature learning & Ingredient and food recognition & - & 1, 2\\
\cite{hannan2013automated} & Hand-crafted features & Bin level detection & Truck & 3\\
\cite{rad2017computer} & Feature learning & Street waste recognition & Truck & 1, 2\\
\cite{zhang2019urban} & Feature learning & Street cleanliness assessment & Truck & 1, 2\\
\cite{mittal2016spotgarbage} & Feature learning & Waste detection & Smartphone & 1, 2\\
\cite{gao2019musefood} & Feature learning & Food volume estimation & Smartphone & 3\\
\cite{pouladzadeh2017mobile} & Feature learning & Food recognition & Smartphone & 2\\
\cite{mao2021recycling} & Feature learning & Waste recognition & - & 1, 2\\
\cite{vo2019novel} & Feature learning & Waste recognition & - & 1, 2\\
\cite{jahanian2019see} & Feature learning & Waste recognition & Robot & 2\\
\hline 
\end{tabular}
\end{table}
\revision{The selected works (32 in total) are also organized in the contingency matrix of  Fig. \ref{fig:ContingencyMatrix} which has the conference/journal of the paper specified along the x-axis and the keywords extracted from the paper abstracts along the y-axis. The matrix shows that ``convolutional neural network'' and ``material recognition'' have the highest frequencies (9 and 10, respectively); the IEEE Conference on Computer Vision and Pattern Recognition (CVPR) has the highest frequency among the x-axis terms. Note that in the field of Computer Vision and Machine Learning, conferences such as CVPR or ICCV have higher or equivalent quality outputs compared to journals, e.g. IEEE Transactions on Pattern Analysis and Machine Intelligence or IEEE Transactions on Image Processing (with frequencies 2 and 1, respectively). This is the opposite of other research fields, in which the journals have higher impact than conferences. The keyword ``trash classification'', which has a frequency of 4, has the highest number of mentions in the selected papers published in IEEE Access (2 papers in total). The term ``optimized densenet121'' has the highest number of mentions in Resources, Conservation and Recycling, which is a journal more focused on the application of Computer Vision; indeed, DenseNet-121 was previously proposed at CVPR with no regard for the specific domain of application \cite{huang2017densely}. Finally, it should be noted that the majority of the journals/conferences of the 32 selected works are more oriented to Computer Vision research than to Circular Economy. This is because, as in the title, the review emphasizes the foundational aspects of MMUs. However, it does so by covering these foundational aspects within the context of the target area of application, that is, Circular Economy. The main authors' goal is to make a manuscript accessible and of interest for both research communities.}    

\begin{figure}
\centering
\includegraphics[width=0.8\textwidth]{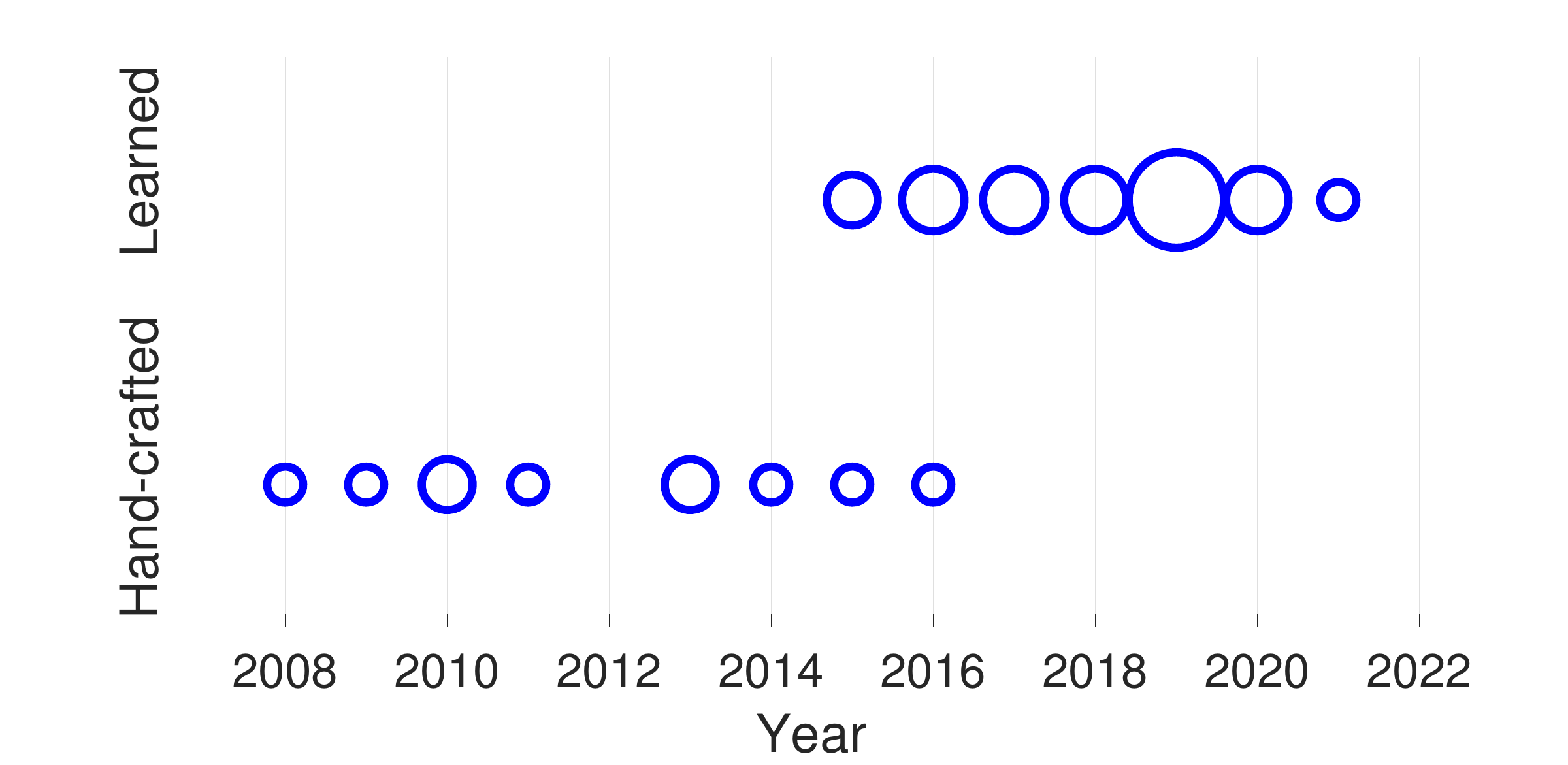}
\caption{\revision{Approach used as a function of the publication year for the selected works listed in Table \ref{Tab:Summary}. The diameters of the circles are proportional to the number of articles with the same approach-year pair.}}
\label{fig:MethodVSyear}
\end{figure}

\revision{Figure \ref{fig:MethodVSyear} shows the approaches used in the selected works as specified in the second column of Table \ref{Tab:Summary} as a function of the year of publication. The diameter of a circle is proportional to the number of articles with the same approach-year pair. It is visible that hand-crafted features are the dominant approach until 2014, whereas the two approaches overlap in 2015 and 2016; after 2016, the dominant approach is feature learning, which is mainly underpinned by deep learning.}  

\section{\revision{Conclusions}}\label{sec:Conclusions}
To improve the management of natural resources, this paper first defined a computer-vision-based complex sensor and then reviewed the works related to its technological foundations. A network of such sensors for wide-area material stock monitoring was also discussed. Accurate estimation of the mass of materials by processing real-world images is the most challenging task, especially if the vision system is implemented on edge devices with limited computing performance such as microcontrollers and smartphones. Five future research directions have been outlined. We hope to reach the interest of \revision{computer vision researchers and} engineers concerned about the human ecological footprint and the interest of \revision{industrial ecologists, circular economists and environmental researchers and engineers} looking at new ways to use the latest advances in computer vision.

\section*{Acknowledgments}
The first author gratefully acknowledges Irish Manufacturing Research (IMR) for the financial support provided for this work. All the authors thank the creators of the datasets in Table \ref{Tab:Datasets} for making them publicly available.

\bibliography{mybibfile}

\begin{thebibliography}{100}
\expandafter\ifx\csname url\endcsname\relax
  \def\url#1{\texttt{#1}}\fi
\expandafter\ifx\csname urlprefix\endcsname\relax\def\urlprefix{URL }\fi
\expandafter\ifx\csname href\endcsname\relax
  \def\href#1#2{#2} \def\path#1{#1}\fi

\bibitem{EllenAndGranta2015}
E.~M. Foundation, G.~Design,
  \href{https://www.ellenmacarthurfoundation.org/assets/downloads/insight/Circularity-Indicators_Project-Overview_May2015.pdf}{{Circularity
  indicators: An approach to measuring circularity (project overview)}} (2015).
\newline\urlprefix\url{https://www.ellenmacarthurfoundation.org/assets/downloads/insight/Circularity-Indicators_Project-Overview_May2015.pdf}

\bibitem{EuropeanCommission2018}
E.~Commission,
  \href{https://eur-lex.europa.eu/legal-content/EN/TXT/PDF/?uri=CELEX:52018DC0773}{{A
  clean planet for all: A European strategic long-term vision for prosperous,
  modern, competitive and climate neutral economy; COM(2018) 773 final},
  brussels} (2018).
\newline\urlprefix\url{https://eur-lex.europa.eu/legal-content/EN/TXT/PDF/?uri=CELEX:52018DC0773}

\bibitem{EuropeanCommission2019}
E.~Commission,
  \href{https://ec.europa.eu/info/sites/default/files/european-green-deal-communication_en.pdf}{{The
  European green deal; COM(2019) 640 final}, brussels} (2019).
\newline\urlprefix\url{https://ec.europa.eu/info/sites/default/files/european-green-deal-communication_en.pdf}

\bibitem{EuropeanCommission2020}
E.~Commission, {Bringing nature back into our lives: EU 2030 biodiversity
  strategy} (2020).

\bibitem{CambridgeEconometrics2018}
C.~Econometrics, Trinomics, ICF,
  \href{https://circulareconomy.europa.eu/platform/sites/default/files/ec_2018_-_impacts_of_circular_economy_policies_on_the_labour_market.pdf}{{Impacts
  of circular economy policies on the labour market: Final report}} (2018).
\newline\urlprefix\url{https://circulareconomy.europa.eu/platform/sites/default/files/ec_2018_-_impacts_of_circular_economy_policies_on_the_labour_market.pdf}

\bibitem{UnitedNations2030Agenda}
U.~Nations, \href{https://sdgs.un.org/2030agenda}{{Transforming our world: The
  2030 agenda for sustainable development}} (2015).
\newline\urlprefix\url{https://sdgs.un.org/2030agenda}

\bibitem{EuropeanParliamentWaste2021}
E.~Parliament,
  \href{https://www.europarl.europa.eu/news/en/headlines/society/20180328STO00751/eu-waste-management-infographic-with-facts-and-figures}{{Waste
  management in the EU: Infographic with facts and figures}; updated on
  03-03-2021} (2021).
\newline\urlprefix\url{https://www.europarl.europa.eu/news/en/headlines/society/20180328STO00751/eu-waste-management-infographic-with-facts-and-figures}

\bibitem{OECDoutlook2018}
OECD,
  \href{https://www.oecd.org/environment/waste/highlights-global-material-resources-outlook-to-2060.pdf}{{Global
  material resources outlook to 2060: Economic drivers and environmental
  consequences (highlights)}} (2018).
\newline\urlprefix\url{https://www.oecd.org/environment/waste/highlights-global-material-resources-outlook-to-2060.pdf}

\bibitem{kaza2018waste}
S.~Kaza, L.~Yao, P.~Bhada-Tata, F.~Van~Woerden, What a waste 2.0: A global
  snapshot of solid waste management to 2050, World Bank Publications, 2018.

\bibitem{adedeji2017leakage}
K.~B. Adedeji, Y.~Hamam, B.~T. Abe, A.~M. Abu-Mahfouz, Leakage detection and
  estimation algorithm for loss reduction in water piping networks, Water
  9~(10) (2017) 773.

\bibitem{hannan2015review}
M.~Hannan, M.~A. Al~Mamun, A.~Hussain, H.~Basri, R.~A. Begum, A review on
  technologies and their usage in solid waste monitoring and management
  systems: {Issues} and challenges, Waste Management 43 (2015) 509--523.

\bibitem{gundupalli2017review}
S.~P. Gundupalli, S.~Hait, A.~Thakur, A review on automated sorting of
  source-separated municipal solid waste for recycling, Waste Management 60
  (2017) 56--74.

\bibitem{shapiro2000computer}
L.~Shapiro, G.~Stockman,
  \href{http://nana.lecturer.pens.ac.id/index_files/referensi/computer_vision/Computer\%20Vision.pdf}{Computer
  vision (march 2000)} (2000).
\newline\urlprefix\url{http://nana.lecturer.pens.ac.id/index_files/referensi/computer_vision/Computer\%20Vision.pdf}

\bibitem{minsky1961steps}
M.~Minsky, Steps toward artificial intelligence, Proceedings of the IRE 49~(1)
  (1961) 8--30.

\bibitem{szeliski2020computer}
R.~Szeliski, \href{https://szeliski.org/Book/}{Computer vision: Algorithms and
  applications; 2nd Edition}, Springer, Draft of September 2020.
\newline\urlprefix\url{https://szeliski.org/Book/}

\bibitem{forsyth2012computer}
D.~A. Forsyth, J.~Ponce,
  \href{https://eclass.teicrete.gr/modules/document/file.php/TM152/Books/Computer\%20Vision\%20-\%20A\%20Modern\%20Approach\%20-\%20D.\%20Forsyth,\%20J.\%20Ponce.pdf}{Computer
  vision: A modern approach; 2nd Edition}, Pearson, 2012.
\newline\urlprefix\url{https://eclass.teicrete.gr/modules/document/file.php/TM152/Books/Computer\%20Vision\%20-\%20A\%20Modern\%20Approach\%20-\%20D.\%20Forsyth,\%20J.\%20Ponce.pdf}

\bibitem{Goodfellow-et-al-2016}
I.~Goodfellow, Y.~Bengio, A.~Courville,
  \href{http://www.deeplearningbook.org}{Deep Learning}, MIT Press, 2016.
\newline\urlprefix\url{http://www.deeplearningbook.org}

\bibitem{zhang2020dive}
A.~Zhang, Z.~C. Lipton, M.~Li, A.~J. Smola, \href{https://d2l.ai/}{Dive into
  deep learning; release 0.15.0}.
\newline\urlprefix\url{https://d2l.ai/}

\bibitem{wang2018deep}
J.~Wang, Y.~Ma, L.~Zhang, R.~X. Gao, D.~Wu, Deep learning for smart
  manufacturing: Methods and applications, Journal of Manufacturing Systems 48
  (2018) 144--156.

\bibitem{csurka2004visual}
G.~Csurka, C.~Dance, L.~Fan, J.~Willamowski, C.~Bray, Visual categorization
  with bags of keypoints, in: Workshop on statistical learning in computer
  vision, ECCV, Vol.~1, Prague, 2004, pp. 1--2.

\bibitem{csurka2006generic}
G.~Csurka, C.~R. Dance, F.~Perronnin, J.~Willamowski, Generic visual
  categorization using weak geometry, in: Toward Category-Level Object
  Recognition, Springer, 2006, pp. 207--224.

\bibitem{mikolajczyk2002affine}
K.~Mikolajczyk, C.~Schmid, An affine invariant interest point detector, in:
  European conference on computer vision, Springer, 2002, pp. 128--142.

\bibitem{lowe1999object}
D.~G. Lowe, Object recognition from local scale-invariant features, in:
  Proceedings of the seventh IEEE international conference on computer vision,
  Vol.~2, IEEE, 1999, pp. 1150--1157.

\bibitem{zhang2007local}
J.~Zhang, M.~Marsza{\l}ek, S.~Lazebnik, C.~Schmid, Local features and kernels
  for classification of texture and object categories: A comprehensive study,
  International journal of computer vision 73~(2) (2007) 213--238.

\bibitem{felzenszwalb2005pictorial}
P.~F. Felzenszwalb, D.~P. Huttenlocher, Pictorial structures for object
  recognition, International journal of computer vision 61~(1) (2005) 55--79.

\bibitem{FergusCourse2009}
R.~Fergus,
  \href{http://people.csail.mit.edu/torralba/shortCourseRLOC/}{Classical
  methods for object recognition}, in: ICCV 2009 Short Course on Recognizing
  and Learning Object Categories, Kyoto, Japan, 2009.
\newline\urlprefix\url{http://people.csail.mit.edu/torralba/shortCourseRLOC/}

\bibitem{oliva2007role}
A.~Oliva, A.~Torralba, The role of context in object recognition, Trends in
  cognitive sciences 11~(12) (2007) 520--527.

\bibitem{sudderth2008describing}
E.~B. Sudderth, A.~Torralba, W.~T. Freeman, A.~S. Willsky, Describing visual
  scenes using transformed objects and parts, International Journal of Computer
  Vision 77~(1-3) (2008) 291--330.

\bibitem{crandall2007composite}
D.~J. Crandall, D.~P. Huttenlocher, Composite models of objects and scenes for
  category recognition, in: 2007 IEEE Conference on Computer Vision and Pattern
  Recognition, IEEE, 2007, pp. 1--8.

\bibitem{muja2014scalable}
M.~Muja, D.~G. Lowe, Scalable nearest neighbor algorithms for high dimensional
  data, IEEE transactions on pattern analysis and machine intelligence 36~(11)
  (2014) 2227--2240.

\bibitem{bishop2006pattern}
C.~M. Bishop, Pattern recognition and machine learning, Springer, 2006.

\bibitem{lampert2009kernel}
C.~H. Lampert, Kernel methods in computer vision, Now Publishers Inc, 2009.

\bibitem{criminisi2013decision}
A.~Criminisi, J.~Shotton, Decision forests for computer vision and medical
  image analysis, Springer, 2013.

\bibitem{lecun2010convolutional}
Y.~LeCun, K.~Kavukcuoglu, C.~Farabet, Convolutional networks and applications
  in vision, in: Proceedings of 2010 IEEE international symposium on circuits
  and systems, IEEE, 2010, pp. 253--256.

\bibitem{MATLABtutorial}
MathWorks,
  \href{https://uk.mathworks.com/help/deeplearning/ug/create-simple-deep-learning-network-for-classification.html}{Create
  simple deep learning network for classification}.
\newline\urlprefix\url{https://uk.mathworks.com/help/deeplearning/ug/create-simple-deep-learning-network-for-classification.html}

\bibitem{ren2015faster}
S.~Ren, K.~He, R.~Girshick, J.~Sun, Faster {R-CNN}: Towards real-time object
  detection with region proposal networks, in: Advances in neural information
  processing systems, 2015, pp. 91--99.

\bibitem{zhao2019object}
Z.-Q. Zhao, P.~Zheng, S.-t. Xu, X.~Wu, Object detection with deep learning: A
  review, IEEE transactions on neural networks and learning systems 30~(11)
  (2019) 3212--3232.

\bibitem{girshick2014rich}
R.~Girshick, J.~Donahue, T.~Darrell, J.~Malik, Rich feature hierarchies for
  accurate object detection and semantic segmentation, in: Proceedings of the
  IEEE conference on computer vision and pattern recognition, 2014, pp.
  580--587.

\bibitem{girshick2015fast}
R.~Girshick, Fast {R-CNN}, in: Proceedings of the IEEE international conference
  on computer vision, 2015, pp. 1440--1448.

\bibitem{he2015spatial}
K.~He, X.~Zhang, S.~Ren, J.~Sun, Spatial pyramid pooling in deep convolutional
  networks for visual recognition, IEEE transactions on pattern analysis and
  machine intelligence 37~(9) (2015) 1904--1916.

\bibitem{long2015fully}
J.~Long, E.~Shelhamer, T.~Darrell, Fully convolutional networks for semantic
  segmentation, in: Proceedings of the IEEE conference on computer vision and
  pattern recognition, 2015, pp. 3431--3440.

\bibitem{dai2016r}
J.~Dai, Y.~Li, K.~He, J.~Sun, {R-FCN}: Object detection via region-based fully
  convolutional networks, in: Advances in neural information processing
  systems, 2016, pp. 379--387.

\bibitem{redmon2016you}
J.~Redmon, S.~Divvala, R.~Girshick, A.~Farhadi, You only look once: Unified,
  real-time object detection, in: Proceedings of the IEEE conference on
  computer vision and pattern recognition, 2016, pp. 779--788.

\bibitem{liu2016ssd}
W.~Liu, D.~Anguelov, D.~Erhan, C.~Szegedy, S.~Reed, C.-Y. Fu, A.~C. Berg, Ssd:
  Single shot multibox detector, in: European conference on computer vision,
  Springer, 2016, pp. 21--37.

\bibitem{voulodimos2018deep}
A.~Voulodimos, N.~Doulamis, A.~Doulamis, E.~Protopapadakis, Deep learning for
  computer vision: A brief review, Computational intelligence and neuroscience
  2018.

\bibitem{mehta2022using}
N.~Mehta, E.~Cunningham, M.~Doherty, P.~Sainsbury, I.~Bolaji, B.~Firoozi-Nejad,
  B.~M. Smyth, Using regional material flow analysis and geospatial mapping to
  support the transition to a circular economy for plastics, Resources,
  Conservation and Recycling 179 (2022) 106085.

\bibitem{millette2019materials}
S.~Millette, E.~Williams, C.~E. Hull, Materials flow analysis in support of
  circular economy development: Plastics in {Trinidad and Tobago}, Resources,
  Conservation and Recycling 150 (2019) 104436.

\bibitem{MFA-Europe}
\relax Joint Research Centre~(JRC),
  \href{https://rmis.jrc.ec.europa.eu/uploads/scoreboard2018/indicators/15._Material_flows_in_the_circular_economy.pdf}{European
  innovation partnership on raw materials: Material flows in the circular
  economy (last access on 28 {February} 2022)}.
\newline\urlprefix\url{https://rmis.jrc.ec.europa.eu/uploads/scoreboard2018/indicators/15._Material_flows_in_the_circular_economy.pdf}

\bibitem{myers2019ystafdb}
R.~J. Myers, B.~K. Reck, T.~Graedel, {YSTAFDB}, a unified database of material
  stocks and flows for sustainability science, Scientific data 6~(1) (2019)
  1--13.

\bibitem{DensityOfMaterials}
A.~I. of~Physics,
  \href{https://web.mit.edu/8.13/8.13c/references-fall/aip/aip-handbook.html#sec1}{{American
  Institute of Physics Handbook, Section 2b}} (1972).
\newline\urlprefix\url{https://web.mit.edu/8.13/8.13c/references-fall/aip/aip-handbook.html#sec1}

\bibitem{sharan2013recognizing}
L.~Sharan, C.~Liu, R.~Rosenholtz, E.~H. Adelson, Recognizing materials using
  perceptually inspired features, International journal of computer vision
  103~(3) (2013) 348--371.

\bibitem{bell2015material}
S.~Bell, P.~Upchurch, N.~Snavely, K.~Bala, Material recognition in the wild
  with the materials in context database, in: Proceedings of the IEEE
  conference on computer vision and pattern recognition, 2015, pp. 3479--3487.

\bibitem{schwartz2019recognizing}
G.~Schwartz, K.~Nishino, Recognizing material properties from images, IEEE
  transactions on pattern analysis and machine intelligence 42~(8) (2019)
  1981--1995.

\bibitem{zocco2021greedy}
F.~Zocco, M.~Maggipinto, G.~A. Susto, S.~McLoone, Lazy {FSCA} for unsupervised
  variable selection, arXiv preprint arXiv:2103.02687.

\bibitem{wang2020multi}
T.~Wang, Y.~Cai, L.~Liang, D.~Ye, A multi-level approach to waste object
  segmentation, Sensors 20~(14) (2020) 3816.

\bibitem{awe2017smart}
O.~Awe, R.~Mengistu, V.~Sreedhar, Smart trash net: Waste localization and
  classification, Final report, {Stanford University}.

\bibitem{li2012recognizing}
W.~Li, M.~Fritz, Recognizing materials from virtual examples, in: European
  Conference on Computer Vision, Springer, 2012, pp. 345--358.

\bibitem{taco2020}
P.~F. Proença, P.~Simões, Taco: Trash annotations in context for litter
  detection, arXiv preprint arXiv:2003.06975.

\bibitem{cimpoi2014describing}
M.~Cimpoi, S.~Maji, I.~Kokkinos, S.~Mohamed, A.~Vedaldi, Describing textures in
  the wild, in: Proceedings of the IEEE Conference on Computer Vision and
  Pattern Recognition, 2014, pp. 3606--3613.

\bibitem{lowe2004distinctive}
D.~G. Lowe, Distinctive image features from scale-invariant keypoints,
  International journal of computer vision 60~(2) (2004) 91--110.

\bibitem{dalal2005histograms}
N.~Dalal, B.~Triggs, Histograms of oriented gradients for human detection, in:
  2005 IEEE computer society conference on computer vision and pattern
  recognition (CVPR'05), Vol.~1, IEEE, 2005, pp. 886--893.

\bibitem{bo2010kernel}
L.~Bo, X.~Ren, D.~Fox, Kernel descriptors for visual recognition, in: Advances
  in neural information processing systems, 2010, pp. 244--252.

\bibitem{hu2011toward}
D.~Hu, L.~Bo, X.~Ren, Toward robust material recognition for everyday objects,
  in: British Machine Vision Conference, 2011, pp. 48.1--48.11.

\bibitem{liu2010exploring}
C.~Liu, L.~Sharan, E.~H. Adelson, R.~Rosenholtz, Exploring features in a
  bayesian framework for material recognition, in: 2010 {IEEE} computer society
  conference on computer vision and pattern recognition, IEEE, 2010, pp.
  239--246.

\bibitem{blei2003latent}
D.~M. Blei, A.~Y. Ng, M.~I. Jordan, Latent {Dirichlet} allocation, Journal of
  machine learning research 3~(Jan) (2003) 993--1022.

\bibitem{varma2008statistical}
M.~Varma, A.~Zisserman, A statistical approach to material classification using
  image patch exemplars, IEEE transactions on pattern analysis and machine
  intelligence 31~(11) (2008) 2032--2047.

\bibitem{varma2005statistical}
M.~Varma, A.~Zisserman, A statistical approach to texture classification from
  single images, International journal of computer vision 62~(1-2) (2005)
  61--81.

\bibitem{chu2018multilayer}
Y.~Chu, C.~Huang, X.~Xie, B.~Tan, S.~Kamal, X.~Xiong, Multilayer hybrid
  deep-learning method for waste classification and recycling, Computational
  Intelligence and Neuroscience 2018, Article ID 5060857.

\bibitem{bircanouglu2018recyclenet}
C.~Bircano{\u{g}}lu, M.~Atay, F.~Be{\c{s}}er, {\"O}.~Gen{\c{c}}, M.~A.
  K{\i}zrak, Recyclenet: Intelligent waste sorting using deep neural networks,
  in: 2018 Innovations in Intelligent Systems and Applications (INISTA), IEEE,
  2018, pp. 1--7.

\bibitem{mao2021recycling}
W.-L. Mao, W.-C. Chen, C.-T. Wang, Y.-H. Lin, Recycling waste classification
  using optimized convolutional neural network, Resources, Conservation and
  Recycling 164 (2021) 105132.

\bibitem{vo2019novel}
A.~H. Vo, H.~Son, M.~T. Vo, T.~Le, A novel framework for trash classification
  using deep transfer learning, IEEE Access 7 (2019) 178631--178639.

\bibitem{jahanian2019see}
A.~Jahanian, Q.~H. Le, K.~Youcef-Toumi, D.~Tsetserukou, See the e-waste!
  training visual intelligence to see dense circuit boards for recycling, in:
  Proceedings of the IEEE/CVF Conference on Computer Vision and Pattern
  Recognition (CVPR) Workshops, 2019.

\bibitem{he2017mask}
K.~He, G.~Gkioxari, P.~Doll{\'a}r, R.~Girshick, Mask r-cnn, in: Proceedings of
  the IEEE international conference on computer vision, 2017, pp. 2961--2969.

\bibitem{WEEEandCRMs}
W.~Forum, \href{https://weee-forum.org/projects-campaigns/cewaste/}{{Developing
  a voluntary certification scheme for WEEE treatment}; project commenced in
  november 2018} (2018).
\newline\urlprefix\url{https://weee-forum.org/projects-campaigns/cewaste/}

\bibitem{ibrahim2019contaminet}
K.~Ibrahim, D.~A. Savage, A.~Schnirel, P.~Intrevado, Y.~Interian, Contaminet:
  Detecting contamination in municipal solid waste, arXiv preprint
  arXiv:1911.04583.

\bibitem{lagunas2019similarity}
M.~Lagunas, S.~Malpica, A.~Serrano, E.~Garces, D.~Gutierrez, B.~Masia, A
  similarity measure for material appearance, ACM Transactions on Graphics
  (TOG) 38~(4) (2019) 1--12.

\bibitem{min2019survey}
W.~Min, S.~Jiang, L.~Liu, Y.~Rui, R.~Jain, A survey on food computing, ACM
  Computing Surveys (CSUR) 52~(5) (2019) 1--36.

\bibitem{wu2009fast}
W.~Wu, J.~Yang, Fast food recognition from videos of eating for calorie
  estimation, in: 2009 IEEE International Conference on Multimedia and Expo,
  IEEE, 2009, pp. 1210--1213.

\bibitem{anthimopoulos2014food}
M.~M. Anthimopoulos, L.~Gianola, L.~Scarnato, P.~Diem, S.~G. Mougiakakou, A
  food recognition system for diabetic patients based on an optimized
  bag-of-features model, IEEE journal of biomedical and health informatics
  18~(4) (2014) 1261--1271.

\bibitem{o2011introduction}
S.~O'Hara, B.~A. Draper, Introduction to the bag of features paradigm for image
  classification and retrieval, arXiv preprint arXiv:1101.3354.

\bibitem{yang2010food}
S.~Yang, M.~Chen, D.~Pomerleau, R.~Sukthankar, Food recognition using
  statistics of pairwise local features, in: 2010 IEEE Computer Society
  Conference on Computer Vision and Pattern Recognition, IEEE, 2010, pp.
  2249--2256.

\bibitem{kawano2015foodcam}
Y.~Kawano, K.~Yanai, Foodcam: A real-time food recognition system on a
  smartphone, Multimedia Tools and Applications 74~(14) (2015) 5263--5287.

\bibitem{bay2008speeded}
H.~Bay, A.~Ess, T.~Tuytelaars, L.~Van~Gool, Speeded-up robust features (surf),
  Computer vision and image understanding 110~(3) (2008) 346--359.

\bibitem{perronnin2010improving}
F.~Perronnin, J.~S{\'a}nchez, T.~Mensink, Improving the fisher kernel for
  large-scale image classification, in: European conference on computer vision,
  Springer, 2010, pp. 143--156.

\bibitem{graedel2015criticality}
T.~E. Graedel, E.~Harper, N.~T. Nassar, P.~Nuss, B.~K. Reck, Criticality of
  metals and metalloids, Proceedings of the National Academy of Sciences
  112~(14) (2015) 4257--4262.

\bibitem{CriticalityEU}
E.~Commission,
  \href{https://eur-lex.europa.eu/legal-content/EN/TXT/?uri=CELEX:52020DC0474}{{Critical
  Raw Materials Resilience: Charting a Path towards greater Security and
  Sustainability COM(2020) 474 final}} (2020).
\newline\urlprefix\url{https://eur-lex.europa.eu/legal-content/EN/TXT/?uri=CELEX:52020DC0474}

\bibitem{CriticalityUS}
M.~Humphries,
  \href{https://www.everycrsreport.com/files/20190628_R45810_b3112ce909b130b5d525d2265a62ce8236464664.pdf}{{Congressional
  Research Service, Critical Minerals and U.S. Public Policy}} (2019).
\newline\urlprefix\url{https://www.everycrsreport.com/files/20190628_R45810_b3112ce909b130b5d525d2265a62ce8236464664.pdf}

\bibitem{martinel2018wide}
N.~Martinel, G.~L. Foresti, C.~Micheloni, Wide-slice residual networks for food
  recognition, in: 2018 IEEE Winter Conference on Applications of Computer
  Vision (WACV), IEEE, 2018, pp. 567--576.

\bibitem{lecun1999object}
Y.~LeCun, P.~Haffner, L.~Bottou, Y.~Bengio, Object recognition with
  gradient-based learning, in: Shape, contour and grouping in computer vision,
  Springer, 1999, pp. 319--345.

\bibitem{wu2016learning}
H.~Wu, M.~Merler, R.~Uceda-Sosa, J.~R. Smith, Learning to make better mistakes:
  Semantics-aware visual food recognition, in: Proceedings of the 24th ACM
  international conference on Multimedia, 2016, pp. 172--176.

\bibitem{lu2020knowledge}
Y.~Lu, L.~Luo, D.~Huang, Y.~Wang, L.~Chen, Knowledge transfer in vision
  recognition: A survey, ACM Computing Surveys (CSUR) 53~(2) (2020) 1--35.

\bibitem{sun2019exploring}
J.~Sun, K.~Radecka, Z.~Zilic, Exploring better food detection via transfer
  learning, in: 2019 16th International Conference on Machine Vision
  Applications (MVA), IEEE, 2019, pp. 1--6.

\bibitem{xiao2019deep}
G.~Xiao, Q.~Wu, H.~Chen, D.~Cao, J.~Guo, Z.~Gong, A deep transfer learning
  solution for food material recognition using electronic scales, IEEE
  Transactions on Industrial Informatics 16~(4) (2019) 2290--2300.

\bibitem{shorten2019survey}
C.~Shorten, T.~M. Khoshgoftaar, A survey on image data augmentation for deep
  learning, Journal of Big Data 6~(1) (2019) 60.

\bibitem{lu2016food}
Y.~Lu, Food image recognition by using convolutional neural networks (cnns),
  arXiv preprint arXiv:1612.00983.

\bibitem{OurPaperIFAC2020}
F.~Zocco, S.~McLoone, An adaptive memory multi-batch l-bfgs algorithm for
  neural network training, IFAC-PapersOnLine 53~(2) (2020) 8199--8204.

\bibitem{meyers2015im2calories}
A.~Myers, N.~Johnston, V.~Rathod, A.~Korattikara, A.~Gorban, N.~Silberman,
  S.~Guadarrama, G.~Papandreou, J.~Huang, K.~P. Murphy, Im2calories: Towards an
  automated mobile vision food diary, in: Proceedings of the IEEE International
  Conference on Computer Vision, 2015, pp. 1233--1241.

\bibitem{eigen2015predicting}
D.~Eigen, R.~Fergus, Predicting depth, surface normals and semantic labels with
  a common multi-scale convolutional architecture, in: Proceedings of the IEEE
  international conference on computer vision, 2015, pp. 2650--2658.

\bibitem{yu2018methods}
H.~Yu, Z.~Yang, L.~Tan, Y.~Wang, W.~Sun, M.~Sun, Y.~Tang, Methods and datasets
  on semantic segmentation: A review, Neurocomputing 304 (2018) 82--103.

\bibitem{fang2018single}
S.~Fang, Z.~Shao, R.~Mao, C.~Fu, E.~J. Delp, F.~Zhu, D.~A. Kerr, C.~J. Boushey,
  Single-view food portion estimation: Learning image-to-energy mappings using
  generative adversarial networks, in: 2018 25th IEEE International Conference
  on Image Processing (ICIP), IEEE, 2018, pp. 251--255.

\bibitem{goodfellow2014generative}
I.~Goodfellow, J.~Pouget-Abadie, M.~Mirza, B.~Xu, D.~Warde-Farley, S.~Ozair,
  A.~Courville, Y.~Bengio, Generative adversarial nets, in: Advances in neural
  information processing systems, 2014, pp. 2672--2680.

\bibitem{dehais2016two}
J.~Dehais, M.~Anthimopoulos, S.~Shevchik, S.~Mougiakakou, Two-view 3d
  reconstruction for food volume estimation, IEEE transactions on multimedia
  19~(5) (2016) 1090--1099.

\bibitem{HowItsMadeWikipedia}
Wikipedia, \href{https://en.wikipedia.org/wiki/How_It%27s_Made}{How it's made}.
\newline\urlprefix\url{https://en.wikipedia.org/wiki/How_It%27s_Made}

\bibitem{frohs2021book}
W.~Frohs, H.~Jaeger, Industrial carbon and graphite materials: {Raw} materials,
  production and applications, John Wiley \& Sons, February 2021.

\bibitem{dick2014raw}
J.~S. Dick, C.~P. Rader, Raw materials supply chain for rubber products:
  {Overview} of the global use of raw materials, polymers, compounding
  ingredients, and chemical intermediates, Carl Hanser Verlag GmbH Co KG, 2014.

\bibitem{albrecht2006nonwoven}
W.~Albrecht, H.~Fuchs, W.~Kittelmann, Nonwoven fabrics: Raw materials,
  manufacture, applications, characteristics, testing processes, John Wiley \&
  Sons, 2006.

\bibitem{bleicher2020material}
A.~Bleicher, A.~Pehlken, The material basis of energy transitions, Elsevier,
  2020, pp. 1--256.

\bibitem{gupta2015advanced}
K.~Gupta, N.~Gupta, Advanced electrical and electronics materials: Processes
  and applications, John Wiley \& Sons, 2015.

\bibitem{cui2016printed}
Z.~Cui, Printed electronics: Materials, technologies and applications, John
  Wiley \& Sons, 2016.

\bibitem{bajpai2018biermann}
P.~Bajpai, Biermann's handbook of pulp and paper: Raw material and pulp making.
  Volume 1, 3rd edition, Elsevier, 2018.

\bibitem{brydson1999plastics}
J.~A. Brydson, Plastics materials, Elsevier, 1999.

\bibitem{davies2012materials}
G.~Davies, Materials for automobile bodies, Butterworth-Heinemann, 2012.

\bibitem{allwood2012sustainable}
J.~M. Allwood, J.~M. Cullen, M.~A. Carruth, D.~R. Cooper, M.~McBrien, R.~L.
  Milford, M.~C. Moynihan, A.~C. Patel, Sustainable materials: With both eyes
  open, UIT Cambridge Limited, 2012.

\bibitem{allwood2015sustainable}
J.~M. Allwood, J.~M. Cullen, Sustainable materials without the hot air: Making
  buildings, vehicles and products efficiently and with less new material, UIT
  Cambridge Limited, 2015.

\bibitem{piergiovanni2016food}
L.~Piergiovanni, S.~Limbo, Food packaging materials, Springer, 2016.

\bibitem{duggal2017building}
S.~K. Duggal, Building materials, Routledge, 2017.

\bibitem{mouritz2012introduction}
A.~P. Mouritz, Introduction to aerospace materials, Elsevier, 2012.

\bibitem{REEdefinition}
B.~G. Survey,
  \href{https://www2.bgs.ac.uk/mineralsuk/download/mineralProfiles/rare_earth_elements_profile.pdf?_ga=2.198248428.236784483.1617700419-1525726033.1617700419}{Rare
  earth elements} (2011).
\newline\urlprefix\url{https://www2.bgs.ac.uk/mineralsuk/download/mineralProfiles/rare_earth_elements_profile.pdf?_ga=2.198248428.236784483.1617700419-1525726033.1617700419}

\bibitem{lahtela2019novel}
V.~Lahtela, S.~Virolainen, A.~Uwaoma, M.~Kallioinen, T.~K{\"a}rki, T.~Sainio,
  Novel mechanical pre-treatment methods for effective indium recovery from
  end-of-life liquid-crystal display panels, Journal of Cleaner Production 230
  (2019) 580--591.

\bibitem{rabah2008recyclables}
M.~A. Rabah, Recyclables recovery of europium and yttrium metals and some salts
  from spent fluorescent lamps, Waste management 28~(2) (2008) 318--325.

\bibitem{resende2010study}
L.~V. Resende, C.~A. Morais, Study of the recovery of rare earth elements from
  computer monitor scraps--leaching experiments, Minerals Engineering 23~(3)
  (2010) 277--280.

\bibitem{kim2018metal}
Y.~Kim, H.~Seo, Y.~Roh, Metal recovery from the mobile phone waste by chemical
  and biological treatments, Minerals 8~(1) (2018) 8.

\bibitem{thiebaud2018our}
E.~Thi{\'e}baud, L.~M. Hilty, M.~Schluep, H.~W. B{\"o}ni, M.~Faulstich, Where
  do our resources go? indium, neodymium, and gold flows connected to the use
  of electronic equipment in switzerland, Sustainability 10~(8) (2018) 2658.

\bibitem{fishman2018implications}
T.~Fishman, R.~J. Myers, O.~Rios, T.~Graedel, Implications of emerging vehicle
  technologies on rare earth supply and demand in the {United States},
  Resources 7~(1) (2018) 9.

\bibitem{xiu2010materials}
F.-R. Xiu, F.-S. Zhang, Materials recovery from waste printed circuit boards by
  supercritical methanol, Journal of hazardous materials 178~(1-3) (2010)
  628--634.

\bibitem{icsildar2018electronic}
A.~I{\c{s}}{\i}ldar, E.~R. Rene, E.~D. van Hullebusch, P.~N. Lens, Electronic
  waste as a secondary source of critical metals: Management and recovery
  technologies, Resources, Conservation and Recycling 135 (2018) 296--312.

\bibitem{hageluken2006improving}
C.~Hageluken, Improving metal returns and eco-efficiency in electronics
  recycling-a holistic approach for interface optimisation between
  pre-processing and integrated metals smelting and refining, in: Proceedings
  of the 2006 IEEE International Symposium on Electronics and the Environment,
  2006., IEEE, 2006, pp. 218--223.

\bibitem{yoo2009enrichment}
J.-M. Yoo, J.~Jeong, K.~Yoo, J.-c. Lee, W.~Kim, Enrichment of the metallic
  components from waste printed circuit boards by a mechanical separation
  process using a stamp mill, Waste management 29~(3) (2009) 1132--1137.

\bibitem{icsildar2016two}
A.~I{\c{s}}{\i}ldar, J.~van~de Vossenberg, E.~R. Rene, E.~D. van Hullebusch,
  P.~N. Lens, Two-step bioleaching of copper and gold from discarded printed
  circuit boards (pcb), Waste Management 57 (2016) 149--157.

\bibitem{chen2016deep}
J.~Chen, C.-W. Ngo, Deep-based ingredient recognition for cooking recipe
  retrieval, in: Proceedings of the 24th ACM international conference on
  Multimedia, 2016, pp. 32--41.

\bibitem{simonyan2014very}
K.~Simonyan, A.~Zisserman, Very deep convolutional networks for large-scale
  image recognition, in: 3rd International Conference on Learning
  Representations (ICLR), 2015.

\bibitem{chen2020study}
J.~Chen, B.~Zhu, C.-W. Ngo, T.-S. Chua, Y.-G. Jiang, A study of multi-task and
  region-wise deep learning for food ingredient recognition, IEEE Transactions
  on Image Processing 30 (2020) 1514--1526.

\bibitem{borutzky2009bond}
W.~Borutzky, Bond graph methodology: {Development} and analysis of
  multidisciplinary dynamic system models, Springer Science \& Business Media,
  2009.

\bibitem{stoianov2008sensor}
I.~Stoianov, L.~Nachman, A.~Whittle, S.~Madden, R.~Kling, Sensor networks for
  monitoring water supply and sewer systems: {Lessons from Boston}, in: Water
  Distribution Systems Analysis Symposium 2006, 2008, pp. 1--17.

\bibitem{hannan2011radio}
M.~Hannan, M.~Arebey, R.~A. Begum, H.~Basri, Radio frequency identification
  (rfid) and communication technologies for solid waste bin and truck
  monitoring system, Waste management 31~(12) (2011) 2406--2413.

\bibitem{arebey2012solid}
M.~Arebey, M.~Hannan, R.~Begum, H.~Basri, Solid waste bin level detection using
  gray level co-occurrence matrix feature extraction approach, Journal of
  environmental management 104 (2012) 9--18.

\bibitem{hannan2013automated}
M.~Hannan, M.~Arebey, R.~A. Begum, A.~Mustafa, H.~Basri, An automated solid
  waste bin level detection system using {Gabor} wavelet filters and
  multi-layer perception, Resources, Conservation and Recycling 72 (2013)
  33--42.

\bibitem{rad2017computer}
M.~S. Rad, A.~von Kaenel, A.~Droux, F.~Tieche, N.~Ouerhani, H.~K. Ekenel, J.-P.
  Thiran, A computer vision system to localize and classify wastes on the
  streets, in: International Conference on computer vision systems, Springer,
  2017, pp. 195--204.

\bibitem{zhang2019urban}
P.~Zhang, Q.~Zhao, J.~Gao, W.~Li, J.~Lu, Urban street cleanliness assessment
  using mobile edge computing and deep learning, IEEE Access 7 (2019)
  63550--63563.

\bibitem{soni2017smart}
G.~Soni, S.~Kandasamy, Smart garbage bin systems--{A} comprehensive survey, in:
  International Conference on Intelligent Information Technologies, Springer,
  2017, pp. 194--206.

\bibitem{lukka2014zenrobotics}
T.~J. Lukka, T.~Tossavainen, J.~V. Kujala, T.~Raiko, Zenrobotics
  recycler--robotic sorting using machine learning, in: Proceedings of the
  International Conference on Sensor-Based Sorting (SBS), 2014, pp. 1--8.

\bibitem{laszlo2019sorting}
R.~Laszlo, R.~Holonec, R.~Cop{\^\i}ndean, F.~Dragan, Sorting system for e-waste
  recycling using contour vision sensors, in: 2019 8th International Conference
  on Modern Power Systems (MPS), IEEE, 2019, pp. 1--4.

\bibitem{blasco2009development}
J.~Blasco, S.~Cubero, J.~G{\'o}mez-Sanch{\'\i}s, P.~Mira, E.~Molt{\'o},
  Development of a machine for the automatic sorting of pomegranate (punica
  granatum) arils based on computer vision, Journal of food engineering 90~(1)
  (2009) 27--34.

\bibitem{pervsak2020vision}
T.~Per{\v{s}}ak, B.~Viltu{\v{z}}nik, J.~Hernavs, S.~Klan{\v{c}}nik,
  Vision-based sorting systems for transparent plastic granulate, Applied
  Sciences 10~(12) (2020) 4269.

\bibitem{seredkin2019development}
A.~Seredkin, M.~Tokarev, I.~Plohih, O.~Gobyzov, D.~Markovich, Development of a
  method of detection and classification of waste objects on a conveyor for a
  robotic sorting system, in: Journal of Physics: Conference Series, Vol. 1359,
  IOP Publishing, 2019, p. 012127.

\bibitem{tessier2007machine}
J.~Tessier, C.~Duchesne, G.~Bartolacci, A machine vision approach to on-line
  estimation of run-of-mine ore composition on conveyor belts, Minerals
  Engineering 20~(12) (2007) 1129--1144.

\bibitem{kumar2018material}
N.~Kumar, M.~G. Garcia, K.~Tyagi, et~al., Material sorting using a vision
  system, uS Patent App. 15/963,755 (Aug.~30 2018).

\bibitem{mittal2016spotgarbage}
G.~Mittal, K.~B. Yagnik, M.~Garg, N.~C. Krishnan, Spotgarbage: Smartphone app
  to detect garbage using deep learning, in: Proceedings of the 2016 ACM
  International Joint Conference on Pervasive and Ubiquitous Computing, 2016,
  pp. 940--945.

\bibitem{krizhevsky2012imagenet}
A.~Krizhevsky, I.~Sutskever, G.~E. Hinton, Imagenet classification with deep
  convolutional neural networks, in: Advances in neural information processing
  systems, 2012, pp. 1097--1105.

\bibitem{gao2019musefood}
J.~Gao, W.~Tan, L.~Ma, Y.~Wang, W.~Tang, Musefood: Multi-sensor-based food
  volume estimation on smartphones, in: 2019 IEEE SmartWorld, Ubiquitous
  Intelligence \& Computing, Advanced \& Trusted Computing, Scalable Computing
  \& Communications, Cloud \& Big Data Computing, Internet of People and Smart
  City Innovation (SmartWorld/SCALCOM/UIC/ATC/CBDCom/IOP/SCI), IEEE, 2019, pp.
  899--906.

\bibitem{pouladzadeh2017mobile}
P.~Pouladzadeh, S.~Shirmohammadi, Mobile multi-food recognition using deep
  learning, ACM Transactions on Multimedia Computing, Communications, and
  Applications (TOMM) 13~(3s) (2017) 1--21.

\bibitem{MAT-DLtoolbox}
MathWorks, \href{https://uk.mathworks.com/products/deep-learning.html#net}{Deep
  learning toolbox: Design, train, and analyze deep learning networks}.
\newline\urlprefix\url{https://uk.mathworks.com/products/deep-learning.html#net}

\bibitem{stevens2020deep}
E.~Stevens, L.~Antiga, T.~Viehmann, Deep learning with PyTorch, Manning
  Publications, 2020.

\bibitem{TFtutorial}
TensorFlow, \href{https://www.tensorflow.org/tutorials}{Tutorials}.
\newline\urlprefix\url{https://www.tensorflow.org/tutorials}

\bibitem{fei2006one}
L.~Fei-Fei, R.~Fergus, P.~Perona, One-shot learning of object categories, IEEE
  transactions on pattern analysis and machine intelligence 28~(4) (2006)
  594--611.

\bibitem{griffin2007caltech}
G.~Griffin, A.~Holub, P.~Perona, The caltech-256: Caltech technical report, vol
  7694 (2007) 3.

\bibitem{nene1996object}
S.~A. Nene, S.~K. Nayar, H.~Murase, Columbia object image library (coil-100).

\bibitem{lin2014microsoft}
T.-Y. Lin, M.~Maire, S.~Belongie, J.~Hays, P.~Perona, D.~Ramanan,
  P.~Doll{\'a}r, C.~L. Zitnick, Microsoft coco: Common objects in context, in:
  European conference on computer vision, Springer, 2014, pp. 740--755.

\bibitem{zhou2017scene}
B.~Zhou, H.~Zhao, X.~Puig, S.~Fidler, A.~Barriuso, A.~Torralba, Scene parsing
  through {ADE20K} dataset, in: Proceedings of the IEEE Conference on Computer
  Vision and Pattern Recognition, 2017.

\bibitem{OpenImages2}
I.~Krasin, T.~Duerig, N.~Alldrin, V.~Ferrari, S.~Abu-El-Haija, A.~Kuznetsova,
  H.~Rom, J.~Uijlings, S.~Popov, S.~Kamali, M.~Malloci, J.~Pont-Tuset, A.~Veit,
  S.~Belongie, V.~Gomes, A.~Gupta, C.~Sun, G.~Chechik, D.~Cai, Z.~Feng,
  D.~Narayanan, K.~Murphy, {OpenImages: A public dataset for large-scale
  multi-label and multi-class image classification.}, Dataset available from
  https://storage.googleapis.com/openimages/web/index.html.

\bibitem{yang2016classification}
M.~Yang, G.~Thung, Classification of trash for recyclability status, CS229
  Project Report 2016.

\bibitem{sharan2009material}
L.~Sharan, R.~Rosenholtz, E.~Adelson, Material perception: {What} can you see
  in a brief glance?, Journal of Vision 9~(8) (2009) 784--784.

\bibitem{dana1999reflectance}
K.~J. Dana, B.~Van~Ginneken, S.~K. Nayar, J.~J. Koenderink, Reflectance and
  texture of real-world surfaces, ACM Transactions On Graphics (TOG) 18~(1)
  (1999) 1--34.

\bibitem{deng2009imagenet}
J.~Deng, W.~Dong, R.~Socher, L.-J. Li, K.~Li, L.~Fei-Fei, {ImageNet}: A
  large-scale hierarchical image database, in: 2009 IEEE conference on computer
  vision and pattern recognition, IEEE, 2009, pp. 248--255.

\bibitem{MAT-pretrainedModels}
MathWorks,
  \href{https://uk.mathworks.com/help/deeplearning/ug/pretrained-convolutional-neural-networks.html}{Pretrained
  deep neural networks}.
\newline\urlprefix\url{https://uk.mathworks.com/help/deeplearning/ug/pretrained-convolutional-neural-networks.html}

\bibitem{PyTorch-pretrainedModels}
PyTorch, \href{https://pytorch.org/vision/stable/models.html}{Models and
  pre-trained weights}.
\newline\urlprefix\url{https://pytorch.org/vision/stable/models.html}

\bibitem{TF-pretrainedModels}
TensorFlow, \href{https://tfhub.dev/}{{Hello. Welcome to TensorFlow Hub.}}
\newline\urlprefix\url{https://tfhub.dev/}

\bibitem{yosinski2014transferable}
J.~Yosinski, J.~Clune, Y.~Bengio, H.~Lipson, How transferable are features in
  deep neural networks?, in: Proceedings of the 27th International Conference
  on Neural Information Processing Systems, 2014, pp. 3320--3328.

\bibitem{rosenstein2005transfer}
M.~T. Rosenstein, Z.~Marx, L.~P. Kaelbling, T.~G. Dietterich, To transfer or
  not to transfer, in: NIPS 2005 workshop on transfer learning, Vol. 898, 2005,
  pp. 1--4.

\bibitem{wang2021generative}
Z.~Wang, Q.~She, T.~E. Ward, Generative adversarial networks in computer
  vision: A survey and taxonomy, ACM Computing Surveys (CSUR) 54~(2) (2021)
  1--38.

\bibitem{frid2018gan}
M.~Frid-Adar, I.~Diamant, E.~Klang, M.~Amitai, J.~Goldberger, H.~Greenspan,
  Gan-based synthetic medical image augmentation for increased cnn performance
  in liver lesion classification, Neurocomputing 321 (2018) 321--331.

\bibitem{chen2019deep}
J.~Chen, X.~Ran, Deep learning with edge computing: A review., Proceedings of
  the IEEE 107~(8) (2019) 1655--1674.

\bibitem{hwang2017cloud}
K.~Hwang, Cloud computing for machine learning and cognitive applications, MIT
  Press, 2017.

\bibitem{AmazonWS}
Amazon, \href{https://aws.amazon.com/?nc2=h_lg}{{Amazon Web Services}}.
\newline\urlprefix\url{https://aws.amazon.com/?nc2=h_lg}

\bibitem{MicrosoftAzure}
Microsoft,
  \href{https://azure.microsoft.com/en-gb/services/machine-learning/}{{Azure
  Machine Learning}}.
\newline\urlprefix\url{https://azure.microsoft.com/en-gb/services/machine-learning/}

\bibitem{GoogleCloud}
Google, \href{https://cloud.google.com/vertex-ai}{{Vertex AI}}.
\newline\urlprefix\url{https://cloud.google.com/vertex-ai}

\bibitem{IBMcloud}
IBM, \href{https://www.ibm.com/cloud/ai}{{AI solutions}}.
\newline\urlprefix\url{https://www.ibm.com/cloud/ai}

\bibitem{satyanarayanan2017emergence}
M.~Satyanarayanan, The emergence of edge computing, Computer 50~(1) (2017)
  30--39.

\bibitem{mehta2021exploring}
N.~Mehta, E.~Cunningham, D.~Roy, A.~Cathcart, M.~Dempster, E.~Berry, B.~M.
  Smyth, Exploring perceptions of environmental professionals, plastic
  processors, students and consumers of bio-based plastics: Informing the
  development of the sector, Sustainable Production and Consumption 26 (2021)
  574--587.

\bibitem{Benchmarking1}
\relax Papers~with Code,
  \href{https://paperswithcode.com/task/image-classification}{{Image
  classification}}.
\newline\urlprefix\url{https://paperswithcode.com/task/image-classification}

\bibitem{Benchmarking2}
Benchmarks.AI, \href{https://benchmarks.ai/}{{Directory of AI benchmarks}}.
\newline\urlprefix\url{https://benchmarks.ai/}

\bibitem{standley2017image2mass}
T.~Standley, O.~Sener, D.~Chen, S.~Savarese, image2mass: Estimating the mass of
  an object from its image, in: Conference on Robot Learning, PMLR, 2017, pp.
  324--333.

\bibitem{huang2017densely}
G.~Huang, Z.~Liu, L.~Van Der~Maaten, K.~Q. Weinberger, Densely connected
  convolutional networks, in: Proceedings of the IEEE conference on computer
  vision and pattern recognition, 2017, pp. 4700--4708.

\end{thebibliography}

\end{document}